\documentclass[10pt]{article}

\usepackage{amssymb,amsmath}
\usepackage{xspace,nicefrac}
\usepackage{graphicx}

\usepackage[all]{xy}
\usepackage[latin1]{inputenc}
\usepackage{subeqnarray}
\usepackage{ifthen}

\newboolean{PNG}
\setboolean{PNG}{true}

\ifthenelse{\boolean{PNG}}
{
}
{
  \usepackage{epsfig}
}

\newcommand{\almostone}{0.97\columnwidth}

\newcommand{\two}{0.49\columnwidth}
\newcommand{\three}{0.32\columnwidth}

\newcommand{\closetoall}{0.90\textwidth}

\newcommand{\buu}{{\bf u}}
\newcommand{\bv}{{\bf v}}
\newcommand{\bV}{{\bf V}}
\newcommand{\bw}{{\bf w}}
\newcommand{\bx}{{\bf x}}
\newcommand{\by}{{\bf y}}

\newcommand{\bxi}{\boldsymbol{\xi}}
\newcommand{\bphi}{\boldsymbol{\phi}}
\newcommand{\bpsi}{\boldsymbol{\psi}}

\newcommand{\Ll}{{\mathcal L}}   

\newcommand{\Dd}{{\mathcal D}}     

\newcommand{\Cc}{\mathcal{C}}
\newcommand{\Hh}{\mathcal{H}}

\newcommand{\Pp}{\mathcal{P}}

\newcommand{\RR}{{\mathbb R}}

\newcommand{\ZZ}{{\mathbb Z}}

\newcommand{\vv}{{\vec{v}}}
\newcommand{\VV}{{\vec{V}}}

\newcommand{\Div}{{\nabla\cdot}}

\newcommand{\LieD}[2]{\Ll_{#1}{#2}}
\newcommand{\LieDF}[2]{\Ll^+_{#1}{#2}}
\newcommand{\LieDB}[2]{\Ll^-_{#1}{#2}}
\newcommand{\LieDFh}[2]{\Ll^{h+}_{#1}{#2}}
\newcommand{\LieDBh}[2]{\Ll^{h-}_{#1}{#2}}

\newtheorem{thm}{Theorem}[section]
\newtheorem{prop}[thm]{Proposition}

\newcommand{\udesc}[2]{\underset{#1}{\underbrace{#2}}}

\newcommand{\etal}{\textit{et al.\/}\xspace}

\newcounter{enumcntr}
  {\begin{list}{\arabic{enumcntr}.~~ }{\usecounter{enumcntr}%
       \setlength{\labelsep}{0pt}\setlength{\leftmargin}{0pt}%
       \setlength{\labelwidth}{0pt}%
       \setlength{\listparindent}{0pt}}}%
  {\end{list}}
\newenvironment{keywords}%
{\begin{flushleft}\textbf{Keywords.~}}
{\end{flushleft}}

\author{Fran\c{c}ois~Lauze, Mads Nielsen\\
DIKU, Institute of Computer Science, University of Copenhagen\\
Universitetsparken 1, 2100 Kbh Ø\\
Denmark\\ 
\texttt{\{francois,madsn\}@diku.dk}}
\begin{document}
%
\title{On Variational Methods for Motion Compensated
  Inpainting\\DIKU Internal Report\\
  }

\date{January 2009\footnote{This paper is an internal report from the Department of Computer Science, University of Copenhagen, January 2009. Minor modifications were performed in 2011 -- mainly typo corrections -- and this notice was added in September 2018 for the arXiv submission.}
}

\maketitle

\begin{abstract}
  We develop in this paper a generic Bayesian framework for the joint
  estimation of motion and recovery of missing data in a damaged video
  sequence. Using standard maximum a posteriori to variational
  formulation rationale, we derive generic minimum energy formulations
  for the estimation of a reconstructed sequence as well as motion
  recovery.  We instantiate these energy formulations and from their
  Euler-Lagrange Equations, we propose a full multiresolution
  algorithms in order to compute good local minimizers for our
  energies and discuss their numerical implementations, focusing on
  the missing data recovery part, i.e. inpainting. Experimental
  results for synthetic as well as real sequences are presented.
  Image sequences and extra material is available at \texttt{http://image.diku.dk/francois/seqinp.php}.
\end{abstract}


\section{Introduction}
Since the birth of the cinema, more than a century ago, and the apparition of video in the
50's, a huge amount of Motion Picture material has been recorded on different types of
media, including celluloid reels, magnetic tapes, hard disks, DVDs, flash disks...

Unfortunately, this type of material is subject to artifacts, already at image
acquisition, or, for instance, degradations of the storage medium, especially for reels
and tapes, caused by bad manipulation, incorrect storage conditions, or simply the
``patina of time''. These degradations can be of extremely many different types.  For
films, some of the problems encountered include, but are far from limited to, image
flicker, noise, dust, real dust or a deterioration caused by the use of incorrect
chemicals when developing a negative, for instance, missing regions and vertical line
scratches, usually due to mechanical failures in the cameras or projectors, degradation of
the chemicals, color fading, and even the degradation/destruction of the support, as it is
the case for triacetate films, where excessive storage temperature and humidity causes the
formation of acetic acid (vinegar), and this process is auto-catalytic: once started, it
cannot be stopped. Nitrate cellulose based films are highly flammable. Many movie theaters
were burnt down because of that, and their storage is critical because of explosion
risks. Video also is not exempt of problems. The Telecine, a device used to convert a film
to video, can introduce artifacts such as distortions or Moir\'e patterns.  Repeated
playbacks may damage the tape, introducing for instance noise, the lost of synchronization
information produces a disturbing jittering effect. The digital medium has also its own
problems, such as the block artifact due to lossy compression methods, and during network
transmission, some blocks can be lost.

Among these degradations, \emph{blotches} -- i.e, following Kokaram (\cite{kokaram:98}
``regions of high contrast that appear at random position in the frame (...)''. In these
regions, the original data is usually entirely lost and this paper focuses on recovering
the missing data. Digital Inpainting is the name that is commonly given to the process of
filling-in regions of missing data in an image, and although the term was first coined for
spatial images in \cite{bertalmio-sapiro-etal:00}, Image Sequence or Video Inpainting is
now widely used. A main difference resides in the fact that information missing in one
frame is supposed to exist in adjacent frames, this is used to restrict the search space,
via interframe motion estimation.

With the increasing computational power of modern computers, more and more sophisticated
algorithms are developed for inpainting of still frames as well as video sequences. They
are often divided into stochastic and deterministic methods, although the two approaches
often share many modeling aspects.

Spatial Image Inpainting techniques were pioneered by Masnou and Morel
\cite{masnou-morel:98}, location where data is missing is treated as an object occluding
the background and a Elastica based variational model for the completion of broken level
lines is proposed. Bertalmio \etal \cite{bertalmio-sapiro-etal:00} proposed a Partial
Differential Equations (PDE) based method for filling-in regions by a sort of smoothness
transport. They proposed then a variational technique for recovery of level lines and
intensity in \cite{ballester-bertalmio-etal:01}. Chan and Shen proposed in
\cite{chan-shen:01} approaches based on Total Variation as well as Elastica minimization.
Using a different, discrete approach, inspired by work on texture generation
\cite{efros-leung:99,efros-freeman:01}, Criminisi \etal proposed a simple and elegant
patch based method in \cite{criminisi-etal:04}.

Apart from the deterministic/non deterministic division, there exists another level of
modeling, One might speak of a low level analysis approach for a direct study of the image
geometry in the spatial case, or a ``simple'' 2D+time model, while an \emph{high level
  analysis} approach will attempt to model/infer key descriptions of the underlying 3D
scene and/or its dynamics using for instance tracking, hypotheses of (quasi)-repetitive
motion , and use them for sequence reconstruction.

In the latter category, Irani and Peleg in \cite{irani-peleg:93} proposed a method that
can among others remove foreground objects from a scene. From a multilayer parametric
motion representation, foreground occluded objects can be detected and removed thanks to
the motion and intensity information available for the other layers.

Extending the spatial exemplar based inpainting work of Criminisi
\etal in, Patwardhan \etal have proposed techniques for video
inpainting, in the case of static background \cite{patwardhan-etal:05}
and extended to relatively simple camera motion
\cite{patwardhan-etal:07}. 

In the other hand, a series of more ``low level'' methods have been able to produced high
quality results. Kokaram, in a series of works \cite{kokaram-etal:95, kokaram:98,
  kokaram:04} explored probabilistic ``2D+1'' based approaches for joint detection and
reconstruction of missing data.  Grossauer in \cite{grossauer:05} used optical flow in
order to detect serious violations which might then be interpreted as blotches, and then
performs a simple transport from neighbor frames. In a similar way, D'Amore \etal
\cite{damore-etal:07}) also developed methods for detection and removal of blotches based
on variational motion estimation.  In a different way, Wexler \etal \cite{wexler-etal:07}
proposed a patch based method for filling in holes, with constrains enforcing spatial en
temporal coherence.  In a recent work, Wang and Mirmehdi \cite{wang-mirmehdi:10} used
spatiotemporal random walks in order to determine best replacement pixels, taking into
account surrounding values.

The work we present here can be seen as a deterministic ``lower level'' approach to the
blotch restoration problem, where we are given a damaged image sequence, as well as the
location of the damages, assumed to have been obtained by specific means. Some recent
automatic approaches provide good results, see for instance Buades \etal
\cite{buades-delon-etal:10} or the one of Wang and Mirmehdi \cite{wang-mirmehdi:09}.  In
the other hand, precise detection of large blotches can be difficult, and semi-automatic
procedures are very often applied.

Starting from a general Bayesian Inference approach, and using standard approaches in
motion recovery, we derive a generic variational formulation for joint estimation of
motion and recovery of missing data, we then instantiate this generic class into several
energy functionals, we propose to solve them via their Euler-Lagrange Equation in a fully
multiscale framework.

Some of these ideas appeared also in the monograph of Aubert and Kornprobst
\cite{aubert-kornprobst:06}, as a contribution from the first author of this paper. We
have also used variations over this framework for Video Deinterlacing in
\cite{keller-lauze-nielsen:08}, temporal super-resolution \cite{keller-lauze-nielsen:10}
and spatial super-resolution \cite{keller-lauze-nielsen:07}. Ideas presented here where
exposed by the authors in a conference paper \cite{lauze-nielsen:04}, and turned to be
similar to the ones of Cocquerez \etal \cite{cocquerez-chanas:03}, but were developed
independently. The connection between these two works was pointed out to the authors by
R. Deriche during a stay of the first author at INRIA Sophia-Antipolis in 2003.

This paper is organized as follows. After having introduced some
notations in the first section, a probabilistic model is discussed in
section \ref{sec:mcbayesian-bayes}, modeling both the purely spatial
behavior for a still frame extracted from a video sequence, as well
the temporal correlation which is expected between the frames,
correlation coming from the \emph{apparent motion}. More precisely,
what the model describes is the joint probability of an image sequence
and a vector field describing apparent motion, knowing that they
should agree with a given degraded image sequence. Using Bayesian
inference, an expression for a posterior probability distribution is
worked out. Then following Mumford in \cite{mumford:94}, a generic
family of variational formulations is deduced for the problems of
simultaneous motion recovery and inpainting, as well as simultaneous
motion recovery, inpainting and denoising. Minimizations for these
variational formulations provide our \emph{variational motion
  compensated inpainting} algorithms.


\section{Bayesian Framework for Inpainting and Motion Recovery}
\label{sec:mcbayesian-bayes}

In this section we introduce a generic probabilistic framework for the
missing data problem, via Bayesian inference. We first provide a
general rationale for writting a posterior probability and via maximum
a posteriori estimation, we derive a generic energy minimization for
the inpainting problem.

\subsection{A Posterior for missing data}
We use discrete settings in order to easily give a sense to the different probability
computations used.  Given the degraded sequence $u_0$, we assume the existence of a
degradation process $\Pp$ which results in $u_0$, this can be the loss of every odd / even
lines in an alternating way, as it is the case in video interlacing, a spatial or
spatiotemporal downsampling or as a less structured collection of missing blocks of
pixels, described by their location $\Omega$ with the sequence's spatial domain $D$.  Our
goal is to reconstruct a sequence $u$ on $D$ and to compute a motion field $\vec{v}$ for
that sequence. We introduce therefore the conditional probability $p(u,\vec{v}|u_0,\Pp)$.
Using The Bayes rule of retrodiction, we can write this conditional probability as a
product of a \emph{likelihood} and a \emph{prior} divided by an \emph{evidence}
\begin{equation}
  \label{eq:thmbayes}
  p(u,\vec{v}|u_0,\Pp)=\frac{p(u_0|u,\vec{v},\Pp)\, 
    p(u,\vec{v}|\Pp)}{p(u_0|\Pp)}.
\end{equation}
The evidence corresponding to the known data plays the role of a normalizing constant and
will be disregarded in the sequel.  We assume that $u_0$ is a degraded version of $u$, with
for instance noise added in the acquisition process or because of aging, and missing data
coming from film abrasion due to a mechanical failure or a bad manipulation. Therefore the
observed degraded image sequence $u_0$ is assumed to depend only on $u$ and not on the
apparent motion $\vv$ -- i.e., occlusion and desocclusion are not considered as
degradations, they ``belong'' to the original sequence. The likelihood term
$p(u_0|u,\vec{v},\Pp)$ is therefore just $p(u_0|u,\Pp)$.  Because of the nature of the
causes for missing data, we can assume the independence of $(u,\vec{v})$ and $\Pp$ and the
prior is simply
$$
p(u,\vec{v}|\Pp) =  p(u,\vec{v}) = p(u|\vec{v})\,p(\vec{v}).
$$
We will now decompose this prior. For that, let's imagine the following common
situation. Using a DVD player, a person is watching a video on a TV set. He/she pushes the
``pause'' button of the DVD player's remote control. There could be motion blur due to
large motion and finite aperture times at image acquisition, but our spectator expects to
see a meaningful still image displayed on the TV screen. When the ``play'' button is
pushed again, the animation resumes, and this person expects to see an animation which is
coherent with the content of the still, at least for a sufficiently small amount of time,
this coherence corresponding of course to the \emph{apparent motion}. This leads us to
assume that $p(u|\vec{v})$ has the form $p(u_s,u_t|\vec{v})$ where $u_s$ and $u_t$ denote
local spatial (still frame) and temporal (animation) distributions for $u$, and we factor
it as
\begin{equation}
  \label{eq:cond_probs}
  p(u_s,u_t|\vec{v}) = p(u_t|u_s,\vec{v})\,p(u_s|\vec{v}).
\end{equation}
As we see, the first term of this factorization is a direct translation of this expected
temporal coherence. For sake of simplicity, we assume the independence of $u_s$ and
$\vec{v}$. This is not necessarily true, as motion edges and image edges have a tendency
to be correlated, a fact exploited by several motion recovery algorithms, starting with
the work of Nagel and Enkelmann (see for instance
\cite{nagel-enkelmann:86,alvarez-weickert-etal:00b,alvarez-deriche-etal:02b}).  On the
other hand, many optical flow algorithms do not use this potential dependency and provide
very accurate motion estimations, in fact often better as approaches that take explicitly
into account spatial edges have a tendency to produce over-segmented flows (see
\cite{memin:03} for instance).  Putting all these elements together, we finally obtain the
following posterior distribution
\begin{equation}
  \label{eq:bayesian_formulation}
  p(u,\vec{v}|u_0,\Pp) \propto  
  \udesc{P_1}{p(u_0|u,\Pp)}\,\, 
  \udesc{P_2}{p(u_s)}\,\,
  \udesc{P_3}{p(u_t|u_s,\vec{v})}\,\,
  \udesc{P_4}{p(\vec{v})}
\end{equation}
where $P_1$ is the likelihood of $u$, $P_2$ is the spatial prior for the sequence, $P_4$
is the motion prior, and $P_3$ is a coupling term that acts both as a temporal prior for
the sequence and a likelihood for the motion -- as is the gray level constancy assumption
along apparent motion trajectories seen from either image or motion point of view.


\section{Variational Formulation}
\label{sec:mcbayes-var}

Among others, a standard way to compute a pair $(\bar{u},\vec{\bar{v}})$
from~(\ref{eq:bayesian_formulation}) is to seek for the \emph{maximum a posteriori} (MAP)
of this expression. The kind of probability functions which are normally used are
\emph{Gibbsian}, i.e. of the form $(1/Z)e^{-E}$ where $E$ is expressed as a sum over
cliques, (\cite{geman-geman:84}) and stochastic algorithms can be used, as for instance
Markov Random Field techniques.

From this point we will assume that the degradation process is given as a known missing
data locus $\Omega\subset D$, the spatiotemporal domain of the sequence, and the
reconstruction will be though as blotch removal, although much of what is presented below
will still formally be valid for deinterlacing or super-resolution. Before discussing
this, we introduce some notations and concepts that will be used in the sequel.

\subsection{Notations}
\label{sec:not}

The approach for sequence modeling taken in this work follows ideas of ordinary
differential equations and dynamical systems.  The spatial domain of a sequence will will
be taken to be the unit square $D_s := (0,1)^2\in\RR^2$. The spatial coordinates will be
denoted $\bx =(x,y)$, the temporal coordinate by $t$ and spatiotemporal coordinate $r =
(\bx,t)$. Given $A\subset B$ the complement of $A$ in $B$ will be denoted by $B\backslash
A$ or just $A^c$ (when $B$ is clear from the context).  $D_t = [0,T]$, $D_t^- = [0,T)$,
$D_t^+ = (0,1]$. $D = D_s\times D_t$, $D^- := D_s\times D_t^-$, $D^+ := D_s\times D_t^+$.
A sequence will be a map $u:D\to\RR^n$, with typically $n=1$ for gray value sequences and
$n=3$ for color ones.  As used above, $\Omega$ will denote the subset of $D$ of missing
data of a sequence.  We assume that $D$ is made of ``planar particles'' $\bx(t)$ that move
smoothly in $D$, at least locally, out of some codimension 2 regions, with velocities $d\bx/dt =
\vv(\bx(t)) = \vv(\bx,t)$ the instantaneous spatial velocity vector field. To each spatial
velocity field $\vv = (v_1,v_2)^t$, we associate a spatiotemporal velocity field $\VV =
(v^t,1)^t$.

In this setting we consider instantaneous vector fields $\vv:D\to\RR^2$. They will not a
priori be a displacement, but a velocity, i.e. the spatial part of the time-derivative of
a particle trajectory.

Given a smooth vector field $X:D\to \RR^3$ smooth enough, The
directional or Lie derivative of a function $f$ in the direction $X$
at $r$ is simply
$$
\LieD{X}{f}(r) = X(r)\cdot \nabla_3 f_{(r)}
$$ 
($\nabla_3$ denotes the \emph{spatio-temporal gradient})\footnote{when $f$
  is vector-valued, $\nabla_3 f$ can be taken to be $\left(Jf\right)^t$, the
  transpose of the Jacobian matrix of $f$ and $\nabla f\cdot X :=
  Jf\,X$}. Standard results on ordinary differential equations show that $X$
has a \emph{local flow} $c^X$, i.e. a local solution of the differential
equation $\dot{c}^X = X\circ c^X$ and that for $h$ small enough the map
$\theta^X_r$, $h\mapsto c^X(r,h)$ is a local diffeomorphism from an open subset
of $D$ into $D$ (see \cite{gallot-hulin-etal:90} for a detailed exposure).
Then on has
\begin{equation}
\label{eq:lieder1}
\LieD{X} f(r) = \lim_{h\to 0}\frac{f(\theta^X_r(h))-f(r)}{h}.
\end{equation}
It makes clear that if $h > 0$, 
\begin{equation}
\label{eq:liederfwrd}
\LieDFh{X}{f}(r) := \frac{f(\theta^X_r(h)) - f(r)}{h}
\end{equation}
is a forward differences approximation of $\LieD{X}{}$, and it we will use
when developing discretizations for numerical solvers, as well as we will
use its \emph{backward difference} approximation 
\begin{equation}
\label{eq:liederbwrd}
\LieDBh{X}{f}(r) := \frac{f(r) - f(\theta^X_t(-h))}{h}.
\end{equation}
We will often forget the $h$ superscript and only write $\LieDB{X}{}$ and $\LieDF{X}{}$.
Given a ``particle'' $\bx$ at time $t$, under the velocity field $\vv$, its position at
time $t+h$ is given by $\theta^\VV_{(\bx,t)}(h)$, where $\VV$ is as above the
spatio-temporal extension of $\vv$, and this position has the form $(\by,t+h)$ and thus
induces a \emph{spatial displacement field}
\begin{equation}
\label{eq:dispfield}
\bv(\bx,h,t) = \by - \bx,
\end{equation}
obtained by integrating the values of $\vv$ in the interval $[t,t+h]$. We will denote its
spatiotemporal extension by $\bV$.

For a spatial velocity field $\vv$, we will denote, abusively, by $\LieD{\vv}{}$
what should be $\LieD{\VV}{}$, and similarly for $\LieDFh{\bv}{}$ and
$\LieDBh{\bv}{}$.  Denoting by $\nabla$ the \emph{spatial} gradient
operator, one has
$$
\LieD{\vv}{f} = V\cdot\nabla_3f = \vv\cdot\nabla f + f_t.
$$

We will use spatial and spatio-temporal divergence operators,
respectively denoted $\Div$ and $\nabla_3\cdot$, although, we will
generally drop the subscript in the last notation, as the dimension of
the vectorial argument of $\Div\,$ should remove ambiguity. For a map
$f:D\to\RR^k$, $J(f)$ will denote the Jacobian of $f$ with respect to
its spatial variables and for a map $g:D\to\RR$, $\Hh(g)$ will denote its
Hessian with respect to spatial variables.

To end this section, we will denote by $\chi_A$ the characteristic function
of a set A:
$$
\chi_A(x) =
\begin{cases}
  1&\text{ if }x\in A\\
  0&\text{ if }x\not\in A
\end{cases}.
$$

\subsection{From MAP to Minimization}

The MAP problem is first transformed into a discrete energy minimization
problem
\begin{eqnarray}\label{eq:inpaint_denoise}
  E(u,\vec{v}) &=& -\log(p(u,\vec{v}|u_0,\Omega)) = \lambda_1E_1(u) + \lambda_2E_2(u_s)\nonumber\\ 
  &&+\lambda_3E_3(u_s,u_t,\vec{v}) + \lambda_4E_4(\vec{v})
\end{eqnarray}
with $\lambda_iE_i = -\log(P_i)$ ($\lambda_i > 0$), and our goal is thus to find a minimizer
$(\bar{u},\vec{\bar{v}})$ of $E(u,\vec{v})$, $E$ being a function of
the image pixel locations and flow locations through $u$ and
$\vv$. When image and motion field spaces are provided Euclidean
structures, and $E$ is reasonably smooth, a necessary (but not
sufficient) condition for $(\bar{u},\vec{\bar{v}})$ to be a minimizer
is to be a zero of the energy gradients with respect to $u$ and $v$,
gradients that represent the corresponding differentials for the
corresponding Euclidean structures, and denoted by $\nabla_uE$ and
$\nabla_vE$ in the sequel:
\begin{equation}
  \begin{cases}
    \nabla_u E(\bar{u},\vec{\bar{v}}) = 0\\
    \nabla_{\vv} E(\bar{u},\vec{\bar{v}}) = 0.
  \end{cases}
\end{equation}
Note however that expression (\ref{eq:inpaint_denoise}) makes sense only
when we can take the $\log$ of each of the $P_i$ distributions.  An
important situation where this is clearly not the case when the likelihood
term has the form of a Dirac distribution:
\begin{equation}
\label{eq:dirac_u0}
  P(u_0|u) = \delta(R u-u_0) = 
  \begin{cases}
    1\quad\text{ if } R u = u_0 \text{ on } D\backslash\Omega\\
    0\quad\text{ otherwise}
  \end{cases}
\end{equation}
where $R$ is an operator such as a spatial or spatiotemporal blur, or simply
the identity. In that situation, we simply seek for a
$(\bar{u},\vec{\bar{v}})$ satisfying
\begin{equation}
\label{eq:inpaint_only}
  \begin{cases}
    (\bar{u},\vec{\bar{v}}) &=
    \underset{(u,\vec{v})}{\text{Argmin}\,}\lambda_2E_2(u_s) +
    \lambda_3E_3(u_s,u_t,\vec{v}) + \lambda_4E_4(\vec{v})\\
    R u = u_0 &\text{ on } D\backslash\Omega.
  \end{cases}
\end{equation}
In that case, a minimizer $(\bar{u},\vec{\bar{v}})$ should satisfy the
following conditions
\begin{equation}
  \begin{cases}
    \nabla_u E(\bar{u},\vec{\bar{v}}) = 0&\text{ on }\Omega\\
    R\bar{u} = u_0&\text{ on }D\backslash\Omega\\
    \nabla_{\vv} E(\bar{u},\vec{\bar{v}}) = 0&~\\
  \end{cases}
\end{equation}
At this point, the formulation we have obtained is discrete in nature. Using the Bayesian
to variational formulation proposed by Mumford in~\cite{mumford:94}, limiting expressions
for the probability distributions/energy involved give rise to a \emph{continuous} energy
formulation, treated at least at a formal level, and deterministic algorithms can be used
to carry out the minimization problem. The continuous expressions will also be denoted
$E_i$.  The ``game'' is therefore to choose meaningful expressions for each of the $E_i$,
which are also computationally \emph{reasonable}.  Variational 2-dimensional formulations
of inpainting can provide the spatial term $E_2$ and variational optical flow algorithms
can be ``plugged-in'' for terms $E_3$ and $E_4$, and really many of them have been
proposed in the two last decades.  We will extend upon these choices.

In the following paragraphs we argue on the instantiations of the different
terms $E_1(u;u_0)$, $E_2(u)$, $E_3(u,v)$ and $E_4(v)$, they should present a
good trade-off between accuracy of modeling and simplicity in order to
result in computationally tractable algorithms.

\subsection{Image data term}
\label{ssec:data}
The data term provides a measure of deviation between an observed sequence
and a candidate reconstruction. As mentioned above, this term is relevant in
the inpainting-denoising case, and has generally the form
$$
E_1(u;u_0) = \int_{\Omega^c}L(Ru,u_0)\,dx\,dt
$$
For an additive noise, $L(x,y) = \phi(x-y)$, for a multiplicative one, $L(x,y) =
\phi(y/x)$. Noise can be very complicated to apprehend, and generally a simple model is
assumed, such as additive Gaussian white noise with a fixed variance or a more general
Laplace noise, and $\phi(x-y) = |x-y|^p$ for a $p \geq 1$. In this work we have restricted
ourselves to the Gaussian white noise case: $\phi(x-y) = (x-y)^2$ and assume that
$R = {\mathbb I}$, the identity transform -- we ignore blurring, or merely we assume
blurring is part of the sequence to be restored.

\subsection{Spatial regularization terms}
\label{ssec:sreg}
Such a term would generally have the form
$$
E_2(u) = \int_DL(\nabla u,\nabla^2u,\dots)\,dx\,dt
$$
where we recall that $\nabla$ denotes the spatial gradient.
Derivatives of order $> 1$ are rarely used in practice
(although see \cite{lysaker-lundervold-tai:03}). Tikhonov
regularization corresponds to the case $L(w) = \phi(|w|^2)$ with
$\phi(x) = x$, but is to be avoided in general, as it blurs edges. One
of the most popular choice is $\phi(x) = \sqrt x$ which corresponds to
the Total Variation of Rudin, Oscher and Fatemi
\cite{rudin-osher-etal:92}.  Because of its non-differentaibility at
$0$, is is often replaced by $\phi(x) = \sqrt{x + \epsilon^2}$ where
$\epsilon > 0$ is small. Other $\phi$, some non convex, have been
reported to behave very well \cite{aubert-kornprobst:06}.

\subsection{Motion likelihood -- Temporal regularization}
\label{ssec:molitp}
As our little story illustrated it, some form of coherence on the
sequence content must be present. It is expressed as the requirement
that some derived quantity from the sequence is conserved across time,
quantities that depend on scene illumination, the image formation
process, etc\dots\, In the temporal discrete setting, this conservation
can be written, using forward differences along a motion field as
\begin{equation}
  \label{eq:consdiscrete}
\LieDF{\bV}{F(u)}(r) = 0, \quad\forall r
\end{equation}
and in the continuous temporal setting, using the Lie-derivative
\begin{equation}
\label{eq:conscont}
\LieD{\vv}(F(u)) = \vv\cdot\nabla F(u) + \frac{\partial F(u)}{\partial t} = 0.
\end{equation} 
The continuous equation is generally derived as an approximation of
the discrete one. In this work we used a view point a bit different:
integrating along fixed time ranges the continuous equation will
provide the discrete one.

The conservation equations \eqref{eq:consdiscrete}--\eqref{eq:conscont} do not
generally hold perfectly, due to a series of factors: among others, occlusions /
disocclusion. Noise can also be a source of error, although presmoothing can
be built into $F$. $F$ can be vectorial in order to express that several
elementary quantities are expected to be conserved, and will always be
assumed to be a linear differential (or integro-differential) operator.

In order to cope with conservation equation failure, the conservation properties
are enforced in a least square or generalized least square sense via terms
$$
E_3(u,\vv) = \int_D L\left(\LieD{\vv}F(u(r))\right)\,dr.
$$
For an $x$ seen as ''a collection of features'' $ (x_1,\dots,x_k)$ where the
$x_i$'s can themselves be vectors, we will always assume that $L(x) =
\phi(\sum_i|x_i|^2)$ or $L(x) = \sum_i\phi_i(|x_i|^2)$ where $|x_i|$ is the
standard Euclidean norm of $x_i$.

The most commonly used operator $F$ is simply the identity: the intensities should match
along the motion trajectories. Although clearly limited in validity
\cite{verri-poggio:89}, it works well in many cases. Higher order differential operators
have been used, such as the spatial gradient, spatial Laplacian, spatial Hessian
\textit{etc}. These can be potentially attractive to enforce specific properties for
motion recovery. The spatial gradient will prefer local translational motion. The
Laplacian, being rotationally invariant would not penalze rotational motion... However one
must keep in mind that a differential operator of degree $n$ will give rise to terms of
degree up to and including $2n$ in the G\^ateaux derivative of $E_3$ wrt $u$.  For that
reason, we have limited our investigations to the identity and spatial gradients.

\subsection{Motion regularization}
\label{ssec:mreg}

A large amount of work has been devoted to regularizer terms for optical
flow algorithms to accommodate the different situations encountered in
practice, ranging from natural image sequences, medical image sequences,
fluid dynamics data... Having Film/Video restoration as our goal, we focus
here on natural image sequences. They are generally composed of moving
rigid/semi-rigid objects projected on the camera plane, resulting mostly in
locally translational motion.

Thus a regularizing term should enforce this behavior, in order to recover
essentially smooth small varying regions with in general 1D discontinuities
between these regions. We will limit ourselves to terms that depend on the
spatial or spatiotemporal Jacobian of the motion field. More details will be
provided in the next paragraph.

\subsection{Proposed variational formulations}
\label{sec:propvar}

From the discussion above, we propose several actual variational formulations for the
joint motion recovery / image sequence inpainting.  For each formulation, we consider the
inpainting/denoising problem and the pure inpainting formulation.  For inpainting
denoising we place ourselves in the additive Gaussian noise situation discussed in
Subsection \ref{ssec:data}:
\begin{equation}
  \label{eq:idata}
  E_1(u;u_0) = \frac{1}{2}\int_{\Omega^c}(u-u_0)^2\,dr.
\end{equation}
The spatial regularization term used in this study is a regularization of
the Total Variation term:
\begin{equation}
  \label{eq:ireg}
  E_2(u) = \int_D\phi(|\nabla u|^2)\,dr
\end{equation}
where $\phi(x^2) = \sqrt{x^2 + \epsilon^2}$ for a certain $\epsilon > 0$. From now, $\phi$
will always denotes this function. This type of regularization in the spatial case is
known to generate starcaising effects artifacts \cite{nikolova:04}. Experiments in Section
\ref{sec:results} on \emph{Inpainting/Denoising} shows some tendency to it, when the
weight of spatial regularization is relatively important. In pure inpainting, however,
this term merely acts as brightness regularization and the above mentioned artifacts are
much less stronger, on invisible.

We present candidates for remaining two terms in the following two
tables. Table~\ref{tab:intmot} deals with the motion likelihood/temporal
regularization part while Table~\ref{tab:motreg} deals with motion
regularization terms. All these terms are ``borrowed'' from well-known
variational optical flow formulations, $\lambda_3$ and $\lambda_4$ being
some weights $>0$. In the sequel, we will refer to one of these terms as
$E_i^k$ where $E_i^k$ is the $k$-term ($k$-row) of the $E_i$ array ($i=3,4$)
presented below.

\begin{table}[ht]
  \centering
  $$
  \begin{array}{|c|c|}
    \hline
    & E_3(u,v) \\\hline
    1 &\int_{D^-}\phi\left((\LieD{\vv}{u})^2\right)dr \\\hline
    2 &\int_{D^-}\phi
    \left((\LieD{\vv}{u})^2+\gamma|\LieD{\vv}{\nabla u}|^2\right)dr \\\hline 
    3 & \int_{D^-}\phi\left((\LieD{\vv}{u})^2\right)+
    \gamma\phi\left(|\LieD{\vv}{\nabla u}|^2\right)dr \\\hline
  \end{array}
  $$
  \caption{Terms coupling intensity  and motion.}
  \label{tab:intmot}
\end{table}
\begin{table}[hpt]
  \centering
  $$
  \begin{array}{|c|c|}
    \hline
    & E_4(v)  \\\hline
    1 &\int_{D^-}\left[\phi\left(|\nabla \vv_1|^2\right) + 
      \phi\left(|\nabla \vv_2|^2\right)\right]dr \\\hline
    2 &\int_{D^-}\phi
    \left(|\nabla \vv_1|^2 + |\nabla \vv_2|^2\right)dr \\\hline
    3 &\int_{D^-}\phi
    \left(|\nabla_3 \vv_1|^2 + |\nabla_3 \vv_2|^2\right)dr \\\hline
  \end{array}
  $$
   \caption{Terms for motion field regularization}
  \label{tab:motreg}
\end{table}

Terms $E_3^1$ and $E_4^1$ are regularizations of the $1$-norm terms proposed
by Aubert \etal (\cite{aubert-deriche-etal:99,aubert-kornprobst:99}).  
Terms $E_3^2$ and $E_3^3$, in their time discrete version where
$\LieDF{V}{}$ is used instead of $\LieD{v}{}$, have been respectively
proposed by Brox \etal \cite{brox-bruhn-etal:04} (that also proposed
regularizer $E_4^3$) and Bruhn and Weickert \cite{bruhn-weickert:05} and
proved to provide highly accurate flow.  The time-continuous version $E_3^2$
was used in the authors' conference paper
\cite{lauze-nielsen:04}. $E_4^2$ was proposed by Weickert and
Schn{\"o}rr in \cite{weickert-schnorr:01b}.


All the combinations were implemented, although we present only some
of them in our experiments, for obvious space reasons.


We have not yet discussed the fact that image sequences can be
vector-valued. For a sequence $u = (u_1,\dots,u_n):D\to\RR^n$, $n>1$, the
same energy terms are used, where we then set
\begin{equation}
  \label{eq:vecvalmags}
  |\nabla u|^2 = \sum_{i=1}^n|\nabla u_i|^2,\quad \left(\LieD{\vv}{u}\right)^2 =
  \sum_{i=1}^n \left(\LieD{\vv}{u_i}\right)^2,\dots 
\end{equation}


\section{Resolution Methodologies}
\label{sec:algo}

Minimizing the energies presented in the previous section is done via solving their
corresponding flow and image Euler-Lagrange equations. Motion estimation requires normally
a multiresolution / multiscale approach in order to be able to handle large
displacements. We adopt a multiresolution approach here and we propose to interleave the
images inpainting steps with the flow recovery ones at each resolution level.  In this
section we describe in more details this resolution approach and we introduce tools for
computing the image Euler-Lagrange equations associated with our formulations. We then
process to the computation of the different energy gradients, at least for the image
related terms. Indeed, the steps for solving motion recovery are well studied and we
follow the references we have mentioned in the previous section, we won't provide the
details of their derivations.
 
\subsection{Multiresolution framework}
\label{sec:mres}

Motion computation depends critically on the range of the different discrete filters
supports used for the estimation of the differential/finite differences quantities in
the optical flow equation, and/or the smoothness of the input image data. For that reason
multiscale or multiresolution techniques are needed in order to avoid meaningless local
minima.  On the other hand, a good approximation of the motion field at a given resolution
should allow a precise recosntruction of image information at that resolution by motion
compensated diffusion.  Therefore we present here a methodology that aims at producing
algorithms converging to a reasonable minimizer by solving iteratively for the image and
the flow, in a complete \emph{spatial} multiresolution setting. We implicitly build
pyramids for the image sequence and the flow. In the sequel we assume that we have $N+1$
resolution levels, level $N$ being the coarsest, level $0$ being the finest, original
one. We do not require a $\nicefrac12$ spatial scaling factor between consecutive levels.

We are seemingly in presence of an hen and egg problem, since in order to compute a flow
with our methods we need image values, that are a priori unknown inside the degradation
locus, and we need a flow in order to inpaint inside the degradation locus. This will be
handled by using a degenerated form of the inpainting algorithm at coarsest resolution,
that allows us to recreate a starting image sequence $u^N$.  At coarsest resolution, the
interframe motion is assumed to be very small, and we inpaint the sequence by computing a
minimizer of the corresponding energy formulation where we have assumed $\vv = 0$ -- no
motion is present. The non-linearity of the image equation provides a form for motion
adaptivity.

\begin{figure}[h]
   \centering 
\ifthenelse{\boolean{PNG}}
{
  \includegraphics[width=\closetoall]{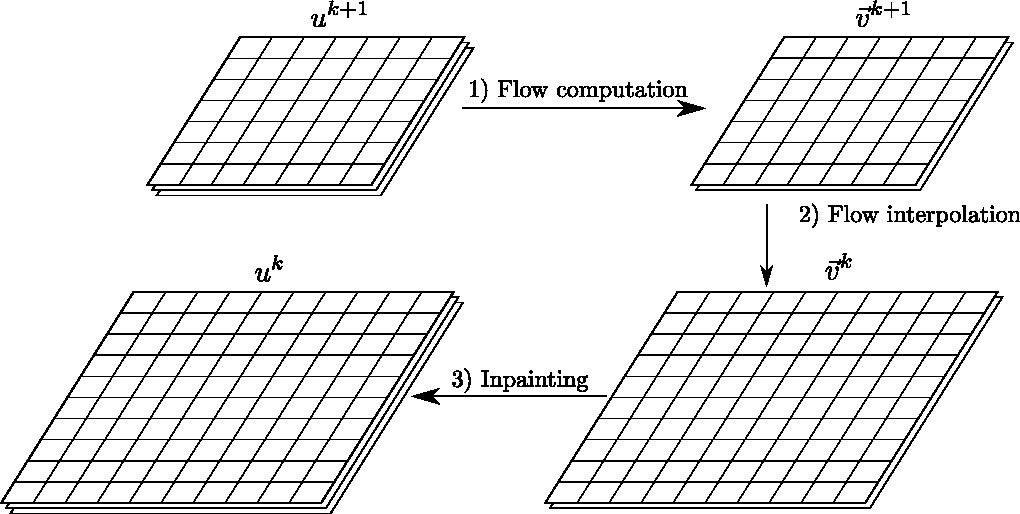}
}
{
  \includegraphics[width=\closetoall]{images/multiresolution.eps}
}
   \caption{Multiresolution inpainting. From a sequence $u^{k+1}$ at
     resolution level $k+1$, 1) the optical flow is computed, 2) it is
     interpolated at resolution level $k$ and 3) used to inpaint the
     sequence at level $k$. This process is then iterated at lower pyramid
     levels.\label{fig:multiresinpaint}}
\end{figure}

Once we have a coarsest resolution inpainted sequence, we can start our
multiresolution process.  What we do is to modify the motion field recovery
algorithm by running the inpainter after the interpolation of the lower
resolution flow, and before updating the flow at that resolution level.
Figure~\ref{fig:multiresinpaint} illustrates a typical iteration. At a given
pyramid level $k+1$, let us assume that we have an inpainted sequence
$u^{k+1}$. It is then used, together with an interpolated version of $\vv^k$,
to compute a motion field at that resolution, to produce $\vv^{k+1}$.  Then at
resolution level $k$, we first interpolate the flow, using for instance a
bilinear or bicubic interpolation, to produce an \emph{intermediary} flow
field $\vv^k_i$.  This flow is used in order to reconstruct the sequence $u^k$
at level $k$.
~\\
A general algorithm (not tied up to our particular formulation) is given in
table~\ref{tab:algo-generic}. As it can be seen from this sketchy algorithm,
the difference between a generic multiresolution optical flow algorithm and
our inpainter are simply the \emph{interleaved} inpainting steps.
\begin{table}[htbp]
  \fbox{ 
    \parbox{\almostone}{
      \begin{enumerate}
      \item Compute $u^N$ by spatial TV and motion adaptive diffusions.
      \item For $k = N-1$ downto 1
        \begin{enumerate}
        \item Compute $\vv^{k+1}$ from $u^{k+1}$
        \item Compute intermediary flow  $\vv^{k}_{i}$ 
          by interpolation of  $\vv^{k+1}$
        \item Inpaint $u^k$ using $\vv^{k}_{i}$
        \end{enumerate}
      \item Compute $\vv^{0}$ from $u^{0}$
      \item Output $(u^0, \vv^{0})$.
      \end{enumerate}
    }}
  \caption{\label{tab:algo-generic}Generic Motion Recovery and Inpainting 
    Algorithm,}
\end{table}
\vspace*{-5mm}

\subsection{Derivation of Euler-Lagrange equations}
\label{sec:eleq}
In this subsection we compute the different energy derivatives with respect
to image variables for the terms proposed in
Subsection~\ref{sec:propvar}. The velocity related derivatives will be
provided without calculation as they have been derived by authors mentioned
in Subsection~\ref{sec:propvar}, at the exception of the new flow smoothing
term. We first provide the general methodology to compute these derivatives
and some elementary results used in the sequel. We restrict ourselve to
scalar valued images and will indicate extension to the vector valued cases.

\subsubsection{Computing functionals differentials}

In order to obtain the Euler-Lagrange equations, we use the technique
of directional derivatives for computing differentials $dE_u$ of
$u\mapsto E(u)$:
\begin{equation}
\label{eq:dirder}
\xi\mapsto dE_u \xi = \ell'(0),\quad \ell(\tau) = E(u+\tau \xi)
\end{equation}
and use of adjunction/ integration by part will allow to transform
$dE_u \xi$, which generally appears as an inner product $\int_\Dd
L_1(u)L_2(\xi)\,dx$ of a term $L_1$ involving derivatives of $u$ and a
term $L_2$ linear in $\xi$ and some of its derivatives on the domain
$\Dd$, into an equivalent expression
$$
dE_u \xi = \int_\Dd F_u \xi\,dx + \int_{\partial \Dd} \xi G_u(\nu)\,ds
$$
where $\partial \Dd$ is the boundary of $\Dd$, $\nu$ its exterior
normal field, and under proper conditions on acceptable deformations
$\xi$ or on $u$ on the boundary $\Dd$ making this boundary integral
vanish, $F_u$ is the \emph{gradient} of $E$ at $u$ for this inner
product.  We will perform computations on each energy part and keep
track of resulting boundary integrals.

We will present integration by part formulas for the spatial gradient
operator $\nabla$ as well as $\LieD{V}{}$. They all result from the
standard integration by part / Stokes formula \cite{evans:98}.
\begin{prop}
  \label{prop:gradadj}
  Let $\Dd \subset \RR^2\times[0,T]$ the domain of a spatiotemporal
  sequence, $\nabla$ be the \emph{spatial} gradient, $\Div$ the
  \emph{spatial} divergence on $D$ and $\nu_s$ the spatial component of the
  exterior unit normal to the boundary $\partial \Dd$ of $\Dd$.
  Let $u:\Dd\to\RR$ and $\vv:\Dd\to\RR^2$ be a $\Cc^1$-vector field  on $\Dd$. Then
  $$
  \int_{\Dd} \nabla u\cdot \vv\,dx = -\int_{\Dd} u\,\Div \vv\,dx +
  \int_{\partial \Dd} u\,\vv\cdot\nu_s\,ds.
  $$
  where $s$ is the area measure on $\partial \Dd$. 
\end{prop}

\begin{prop}
  \label{prop:lieadj}
  Let $\Dd \subset \RR^2\times[0,T]$ the domain of a spatiotemporal sequence, $\varphi$,
  $\psi : \Dd\to \RR^k$, $k\geq 1$, $\vv:\Dd\to\RR^2$ be a $\Cc^1$-vector field and $\VV$
  its ``spatiotemporal'' extension, as introduced in Section \ref{sec:not}. Then one has
  \begin{eqnarray*}
    \int_{\Dd}\left(\LieD{v}\varphi\right)\cdot\psi\,dr &=& 
    -\int_{\Dd}\varphi\,\nabla_3\cdot\left(\psi\otimes V\right)\,dr + 
    \int_{\partial \Dd}\left(\varphi\cdot\psi\right)\left(V\cdot\nu\right)ds\\ 
    &=& 
    -\int_{\Dd}\varphi\cdot\left[\LieD{\vv}\psi + \psi\,\nabla\cdot \vv\right]dr
    + \int_{\partial \Dd}\left(\varphi\cdot\psi\right) \left(\VV\cdot\nu\right)ds
  \end{eqnarray*}
  where $\nu$ is the exterior unit normal to the oriented boundary $\partial \Dd$ of $\Dd$
  and $s$ is the measure on $\partial \Dd$ and $-\cdot-$ is used to denote inner products
  both on $\RR^k$ and $\RR^2$ and $\psi\otimes V =
  \left(\psi_1V^\top,\dots,\psi_kV^\top\right)^\top$ and the divergence is applied componentwise,
  i.e., $\Div_3\left( \psi\otimes V\right) = \left(\Div_3\left(\psi_i V\right)\right)_{i=1\dots k}$.
\end{prop}
The actual velocity field may present discontinuities, as well as functions $\varphi$ and
$\psi$. The simple formula above must then be replaced by a more complex one, as some
terms must be considered as measures \cite{aubert-kornprobst:06}, but we will not consider
it in this work, the problem becoming theoretically very difficult. A practical way to
handle it will be provided by the nonlinearity of the terms considered.

In the following gradient computations, we will for legibility reasons, omit
the different $\lambda_i$ weights from the energy components.
\subsubsection{Data term}

If one wants to use the inpainting/denoising formulation, term
$E_1(u;u_0)$ defined in formula \eqref{eq:idata} must be differentiated with
respect to $u$ and using \eqref{eq:dirder}, one finds immediately that
\begin{equation}
  \label{eq:graduE1}
  dE_{1u}\xi = \int_{\Omega^c}(u-u_{0i})\,\xi\,dr,\quad \nabla_u E_1 =
  \chi_{\Omega^c}(u-u_0).
\end{equation}
When $u_0 = (u_{0i})_{i=1\dots k}$ and $u = (u_{i})_{i=1\dots k}$ have
values in $\RR^k$, equations above are simply replaced by
$$
\nabla_{u_i}E_1 = \chi_{\Omega^c}(u_i - u_{01}),\quad i=1,\dots,k.
$$
\subsubsection{Spatial regularity term}

The derivation of the gradient of the spatial regulariry term $E_2$ given in formula
\eqref{eq:ireg} is rather classical. We recall it however as we are interested in the
boundary terms that come with it. A straightforward computation using directional
derivatives and Prop. \ref{prop:gradadj} leads to
\begin{subeqnarray}
  dE_{2u}\xi &=& -\int_\Dd\xi\Div\left(A\nabla u\right)\,dr\slabel{eq:igrade2}\\
  &+&\int_{\partial \Dd}\xi A\left(\nabla u\cdot\nu_s\right)\,ds\slabel{eq:ie2bc}
\end{subeqnarray}
where we have set
\begin{equation}
  \label{eq:spatialcoef}
  A := A(u) := \phi'(|\nabla u|^2).
\end{equation}
In the pure inpainting case, $\Dd = \Omega$, and we restrict to
deformations $\xi$ that have compact support in $\Omega$, which means
that the boundary term \eqref{eq:ie2bc} vanishes here and the gradient is given by
\begin{equation}
  \label{eq:grade2pi}
  \nabla E_{2u} = -\chi_{\Omega}\Div\left(A\nabla u\right). 
\end{equation}
When $u$ is vector-valued, by using squared-gradient magnitude as
defined in formula \eqref{eq:vecvalmags}, one obtains the $k$ terms
$$
\nabla_{u_i} E_2 = -\chi_\Omega\Div\left(A\nabla u_i\right),\quad i=1,\dots,k
$$
which are coupled via their common diffusivity $A$.
\subsubsection{Temporal prior / motion likelihood term}
\label{sec:tpmlt}

Terms $E_3^i$ couple intensity and motion. We need to compute their gradients with respect
to intensity and with respect to motion, and the novel part is really the intensity
gradient. For that reason we present computations for $E_3^1$ and $E_3^2$ as well as the
result for $E_3^3$.  Here too, integration domains depend on the type of problem we are
interested in, it will be $\Dd = \Omega$ for pure inpainting, and $\Dd = D$ for inpainting
denoising. Gradients with respect to motion fields are provided without any computation
and we won't detail their adaptation in the multiresolution setting, which is standard and
properly explained in the references on optical flow that we provide.
~\\
~\\
\textbf{The term $E_3^1$}.
We start with term $E_3^1$ as defined in Table \ref{tab:intmot}: a
straightforward calculation using formula \eqref{eq:dirder} and Prop~\ref{prop:lieadj} gives
\begin{subeqnarray}
  dE_{3u}^1\xi &=&  -\int_\Dd
  \left[
    \LieD{\vv}{
      \left(B_1\LieD{\vv}{u}\right)
    }
    + B_1\left(\LieD{\vv}{u}\right)\Div\,\vv
  \right]\xi\,dr \slabel{eq:igrade31}
  \\
  &&+
  \int_{\partial\Dd}\xi B_1
  \LieD{\vv}{u}\left(\VV\cdot\nu\right)\,ds\slabel{eq:ie31bc}
\end{subeqnarray}
where we have set
\begin{equation}
\label{eq:b1coef}
B_1 := B_1(u,\vv) := \phi'(|\LieD{\vv}{u}|^2).
\end{equation}
Here too, in the pure inpainting case, $\xi$ has compact support on
$\Omega$ and the boundary integral \eqref{eq:ie31bc} vanishes, the sought gradient is
\begin{eqnarray}
  \label{eq:graduE31}
  \nabla_u E^1_3 &=& -\chi_{\Omega}
  \LieD{\vv}{
    \left[B_1\LieD{\vv}{u}\right]
  }\nonumber\\
 && - \chi_\Omega B_1\left(\LieD{\vv}{u}\right)\Div\,\vv.
\end{eqnarray}
When $u$ is vector valued, $B_1$ couples the different channels via formula
\eqref{eq:vecvalmags} and one obtains $k$ terms:
$$
\nabla_{u_i} E^1_3 = -\chi_\Omega\left(
    \LieD{\vv}{
      \left[B_1\LieD{\vv}{u_i}\right]
    }
    + B_1\left(\LieD{\vv}{u_i}\right)\Div\,\vv
  \right),\quad i=1,\dots,k.
$$
The term  $\nabla_\vv E_3^1$ has been computed by many authors,
see for instance \cite{weickert-schnorr:01b}, and, for the general
vector-valued setting, is given, with our
notations, by the vectorial expression:
$$
\nabla_\vv E_3^1 = B_1\sum_{i=1}^k\left(\LieD{\vv}{u_i}\right)\nabla u_i
$$
and is associated with vanishing Neumann Boundary conditions.
~\\~\\
\textbf{The term $E_3^2$}.  The additional gradient magnitude term in
energy $E_3^2$ induces higher order terms in the
differential. Once the directional derivative is computed, in
order to transform it as a sum of integrals on the domain and a
boundary intergral, we need to apply Prop.~\ref{prop:lieadj} and
Prop. \ref{prop:gradadj}. A careful but nevertheless straightforward
computation provides
\begin{subeqnarray}
  dE_{3u}^2\xi&=&
  -\int_{\Dd}
  \xi
  \left[
    \LieD{\vv}{\left(B_2\LieD{\vv}u\right)} + B_2\left(\LieD{\vv}{u}\right)\Div\,\vv
  \right]
  \,dr \slabel{eq:igrade32-1}\\
  && + \gamma\int_{\Dd}
  \xi
  \Div\left[
    \LieD{\vv}{\left(B_2\LieD{\vv}{\nabla u}\right)} + B_2\left(\LieD{\vv}{\nabla u}\right)\Div\,\vv
  \right]
  \,dr \slabel{eq:igrade32-2}\\
  && - \gamma \int_{\partial\Dd}\xi
  \left[
    \LieD{\vv}{\left(B_2\LieD{\vv}{\nabla u}\right)} + B_2\left(\LieD{\vv}{\nabla u}\right)\Div\,\vv
  \right]\cdot\nu_s\,ds\slabel{eq:ie32bc-1}\\
  && + \int_{\partial \Dd}B_2
  \left[
    \xi\LieD{\vv}{u} + \gamma\nabla\xi\cdot\LieD{\vv}{\nabla u}
  \right]
  (\VV\cdot\nu)\,ds\slabel{eq:ie32bc-2}
\end{subeqnarray}
where we have set
\begin{equation}
\label{eq:b2coef}
B_2 := B_2(u,\vv) :=  \phi'\left(|\LieD{\vv}{u}|^2 +  \gamma|\LieD{\vv}{\nabla u}|^2\right).
\end{equation}
Once again, for the pure inpainting case, boundary integrals
\ref{eq:ie32bc-1} and \ref{eq:ie32bc-2} vanish and the corresponding gradient is
\begin{eqnarray}
  \nabla_u E_3^2 &=& -\chi_{\Omega}
  \left[
    \LieD{\vv}{\left(B_2\LieD{\vv}{u}\right)} - 
    \gamma\Div\left(\LieD{\vv}{\left(B_2\LieD{\vv}{\nabla u}\right)}\right)
  \right] \nonumber\\
  &&- \chi_{\Omega}
  \left[
    B_2 \left(\LieD{\vv}{u}\right)\Div \vv- \gamma\Div\left(B_2\left(\LieD{\vv}{\nabla u}\right)\Div\vv\right)
  \right].
\end{eqnarray}
In the vector valued case, each component is given by the formula above,
replacing $u$ by $u_i$, the components coupling provides from $B_2$. 

The gradient with respect to the motion field $\vv$ of this term is, in the
general vector valued case, given by
$$
\nabla_vE_3^2 = B_2\sum_{i=1}^k\left[\left(\LieD{\vv}{u_i}\right)\nabla u_i +
  \Hh(u_i)\LieD{\vv}{(\nabla u_i)}\right].
$$
~\\
\textbf{The term $E_3^3$}.
Similar computations can be performed for the term $E_3^3$ and similar
complex boundary integral terms appear in the computations.
We set 
\begin{equation}
  \label{eq:b3coef}
  B_3 := B_3(u,\vv) := \phi'(|\LieD{\vv}{\nabla u}|^2)
\end{equation}
and together with $B_1$ that was defined in formula \eqref{eq:b1coef},
we get 
\begin{subeqnarray}
  dE_{3u}^3\xi &=&
  -\int_{\Dd}
  \xi
  \left[
    \LieD{\vv}{\left(B_1\LieD{\vv}u\right)} + B_1\left(\LieD{v}{u}\right)\Div\,v
  \right]
  \,dr \slabel{eq:igrade33-1}\\
  && + \gamma\int_{\Dd}
  \xi
  \Div\left[
    \LieD{\vv}{\left(B_3\LieD{\vv}{\nabla u}\right)} + B_3\left(\LieD{\vv}{\nabla u}\right)\Div\,\vv
  \right]
  \,dr \slabel{eq:igrade33-2}\\
  && - \gamma \int_{\partial\Dd}\xi
  \left[
    \LieD{\vv}{\left(B_3\LieD{\vv}{\nabla u}\right)} + B_3\left(\LieD{\vv}{\nabla u}\right)\Div\,v
  \right]\cdot\nu_s\,ds\slabel{eq:ie33bc-1}\\
  && + \int_{\partial \Dd}
  \left[
    \xi B_1 \LieD{\vv}{u} + \gamma B_3\nabla\xi\cdot\LieD{\vv}{\nabla u}
  \right]
  (\VV\cdot\nu)\,ds\slabel{eq:ie33bc-2}
\end{subeqnarray}
In the pure inpainting case, boundary integrals vanish because of
conditions on $\xi$ and the sought gradient is
\begin{eqnarray}
  \label{eq:pie33grad}
  \nabla E_{3u}^3 &=& -\chi_\Omega\left[
    \LieD{\vv}{(B_1\LieD{\vv}{u})} -\gamma\Div\left(\LieD{\vv}{(B_3\LieD{\vv}{\nabla u})}\right)
    \right]\nonumber\\
    &&-\chi_\Omega\left[
      B_1\left(\LieD{\vv}{u}\right)\Div\,\vv - \gamma\Div\left(B_3\left(\LieD{\vv}{\nabla u}\right)\Div\,\vv\right)
      \right].
\end{eqnarray}

\subsubsection{Boundary terms for minimization with respect to image}
\label{sec:bconds}
So far we have not discussed boundary conditions for the
inpainting/denoising problem where the integration domain is $D$ and
we cannot assume that a deformation direction vanishes along $\partial
D$. The resulting boundary integrals are rather complex. We consider
the case arising from combining spatial regularity term $E_2$ and
temporal regularity term $E_3^1$.  The resulting boundary integral
comes from \eqref{eq:ie2bc} and \eqref{eq:ie31bc}:
$$
\int_{\partial D}\xi\left[\lambda_1\nabla u\cdot\nu_s + \lambda_3 B_1\LieD{\vv}{u}(\VV\cdot \nu)\right]\,ds
$$
and the natural boundary condition is thus to impose that
$$
\lambda_1\nabla u\cdot\nu_s + \lambda_3 B_1\LieD{\vv}{u}(\VV\cdot \nu) = 0.
$$
On domain $D$ exterior normals are respectively $\nu = \pm(1,0,0)^T$,
$\pm(0,1,0)^T$ and $\pm(0,0,1)^T$ corresponding to ``vertical'',
``horizontal'' and ``temporal'' faces of the domain. With these
vectors, the conditions become 
$$
u_x = -\lambda_3v_1B_1\frac{u_yv_2 + u_t}{\lambda_2 +
  \lambda_3v_1B_1},\quad\text{ resp. } u_y =
-\lambda_3v_yB_1\frac{u_xv_1 + u_t}{\lambda_2 +
  \lambda_3v_2B_1},\quad\text{ resp. } u_t = -(u_xv_1 + u_yv_2)
$$
if one uses the expansion $\LieD{\vv}{u} = u_xv_1 + u_yv_2 + u_t$.  However, such an
expansion is problematic numerically. Indeed, it is well known in the optical flow
literature that such an expansion requires sufficient smoothness on $u$ and/or small
motion, and when these requirements are not fulfilled, one has to use approximations such
as the one given in formulas \eqref{eq:liederfwrd} and \eqref{eq:liederbwrd} in the
previous section.

A stronger requirement can be imposed, that both $\LieD{\vv}{u}$ and
$\nabla u\cdot\nu_s$ vanish simultaneously. If it holds, then the
boundary integral vanishes, but imposing such a requirement may lead
to an overdetermined system: for instance, on the ``vertical'' face,
using the above Lie derivative expansion, it becomes
$$
\begin{cases}
  u_x = 0\\
  u_yv_2 + u_t = 0
\end{cases}
$$
and while the first equation expresses the absence of variation across
the boundary, the second is problematic as $u_y$, $u_t$ and $v_2$ are
generally estimated from $\vv$ and $u$ from already known values of the
image.

We will nevertheless assume that $\LieD{\vv}{u}=\nabla u\cdot\nu_s = 0$ as, the use of
standard schemes for computing gradient and the schemes that come from
\eqref{eq:liederfwrd} and \eqref{eq:liederbwrd} for computing $\LieD{\vv}{u}$ provide a
simple and efficient \emph{numerical} treatment of these boundary conditions. These can
also be extended to the terms $E_3²$ and $E_3^3$ by requiring also that $\LieD{\vv}{\nabla
  u} = 0$ on $\partial D$. In fact, without such an assumption, one cannot get rid of
boundary terms appearing in the computations of differentials for $E_3²$ and $E_3^3$,
where not only a deformation $\xi$ appears, but also its spatial gradient $\nabla
\xi$. What makes the difference with the pure inpainting formulation is that it appears
extremely difficult to control what happens at scene boundary. If one could assume null
camera motion and moving objects remaining in the image domain, boundary difficulties
would vanish, but this is in general a much too severe restriction. 

What usually counterbalances the inconsistancies of our choice of boundary conditions is
the reaction to the data term: we are forced to stay reasonably close to the observed data
$u_0$, along the boundary too.

With these assumptions, the three gradient terms $\nabla_uE_3^1$,
$\nabla_uE_3^2$, and $\nabla_uE_3^3$ are easily computed in the
inpainting-denoising formulation and are given by
\begin{eqnarray*}
  \nabla_u E^1_3 &=& -
  \LieD{\vv}{
    \left[B_1\LieD{\vv}{u}\right]
  }\\
  && - B_1\left(\LieD{\vv}{u}\right)Div\,\vv,
\end{eqnarray*}
\begin{eqnarray*}
   \nabla_u E_3^2 &=& -
  \left[
    \LieD{\vv}{\left(B_2\LieD{\vv}{u}\right)} - 
    \gamma\Div\left(\LieD{\vv}{\left(B_2\LieD{\vv}{\nabla u}\right)}\right)
  \right]\\
  &&-
  \left[
    B_2 \left(\LieD{\vv}{u}\right)\Div \vv- \gamma\Div\left(B_2\left(\LieD{\vv}{\nabla u}\right)\Div \vv\right)
  \right],
\end{eqnarray*}
\begin{eqnarray*}
  \nabla E_{3u}^3 &=& -\left[
    \LieD{\vv}{(B_1\LieD{\vv}{u})} -\gamma\Div\left(\LieD{\vv}{(B_3\LieD{\vv}{\nabla u})}\right)
    \right]\\
    &&-\left[
      B_1\left(\LieD{\vv}{u}\right)\Div\,\vv - \gamma\Div\left(B_3\left(\LieD{\vv}{\nabla u}\right)\Div\,\vv\right)
      \right].
\end{eqnarray*}
Two important remarks regarding the structure of these terms (also
valid for their pure inpainting counterpart):
\begin{itemize}
\item First, one observes that these three gradient terms
  $\nabla_uE_3^1$, $\nabla_uE_3^2$, and $\nabla_uE_3^3$ are each
  decomposed in two parts, a part containing a double differentiation
  along the flow field via $\LieD{\vv}{()}$, which corresponds to
  diffusion along a flow line, and a term where $\Div\,\vv$ appears,
  which corrects for the non parallelism of the motion field
  $\vv$. These terms are non linear transport along trajectories of the
  velocity field and are controled by the flow divergence. This means
  that not only the punctual value of $\vv$ must be taken into account
  but also its variations. This is a consequence of
  Prop. \ref{prop:lieadj} and is to be put in parallel with duality
  ideas of Florack \textit{et al.\/} in
  \cite{florack-niessen-etal:98}, but see also
  \cite{wildes-amabile-etal:97,bereziat-herlin-etal:00,memin:03}.

\item  The second remark concerns the assumptions made on conserved quantities in
  order to build the different terms $E_3^i$. The corresponding image
  gradients make clear that these conservations should be enforced for the
  image minimizers of these energies, with intensity diffusion for intensity
  conservation and spatial gradient diffusion for spatial gradient
  conservation.
\end{itemize}

\subsubsection{Flow regularity terms}
\label{sec:flowreg}

As mentioned already, their regularizers $E_4^i$, $i=1,2,3,$ are
borrowed from existing motion recovery algorithms, and their gradient
calculations, with its multiresolution adaptation, have been presented
in the papers already mentioned. We just briefly recall the results.
~\\
~\\
\textbf{The term $E_4^1$}. The two gradient components are
independent, and if one sets
$$
C_1^1 := C_1^1(\vv) := \phi'\left(|\nabla v_1|^2\right),
\quad C_1^2 := C_1^2(\vv) := \phi'\left(|\nabla v_2|^2\right),
$$
one gets
$$
\nabla_\vv E_4^1 = -
\begin{pmatrix}
  \Div\left(C_1^1\nabla v_1\right)\\
  \Div\left(C_1^2\nabla v_2\right)\\
\end{pmatrix}.
$$
~\\
~\\
\textbf{The term $E_4^2$}. The two gradient components are coupled via the
common diffusivity
$$
C_2(\vv) = \phi'\left(|\nabla v_1|^2 + |\nabla v_2|^2\right)
$$
and
$$
\nabla_\vv E_4^2 = -
\begin{pmatrix}
  \Div\left(C_2\nabla v_1\right)\\
  \Div\left(C_2\nabla v_2\right)\\
\end{pmatrix}.
$$
~\\
~\\
\textbf{The term $E_4^3$}. This term is a simple spatio-temporal extension
of the previous one, the new diffusivity is
$$
C_3(\vv) = \phi'\left(|\nabla_3 v_1|^2 + |\nabla_3 v_2|^2\right)
$$
and 
$$
\nabla_\vv E_4^3 = -
\begin{pmatrix}
  \Div\left(C_3\nabla_3 v_1\right)\\
  \Div\left(C_3\nabla_3 v_2\right)\\
\end{pmatrix}
$$
where the divergences are spatio-temporal here since we use the
spatio-temporal gradients of the $v_i$.

We note that these gradients
formulations are naturally associated to Neumann boundary conditions on the
components of $\vv$, the main derivation mechanism being used is given by
Prop. \ref{prop:gradadj}. 

\section{Discretization}
\label{sec:discr}
For optical flow recovery parts, we use discretizations that have been proposed in the
papers where we ``borrowed'' this terms. The discretization of spatial (and
spatio-temporal) divergence terms are quite standard in the literature, see for instance
\cite{chan-shen:02a,aubert-kornprobst:06} and won't be discussed in this paper.

We will thus exclusively concentrate on the part of the inpainting equations that come
from diffusion along flow lines i.e., terms of the type $E_3^i$.  Because, in the numerics
we need to use time discretized sequences, displacement fields must be used instead of
velocity fields, we modify in a natural way our formulations to handle them, with
classical warping techniques. These displacement fields do usually provide subpixel
accuracy, making necessary the use of some form of spatial interpolation, which in turn
make cumbersome the direct development of completely discrete schemes. In order to avoid
that, we use an intermediary formulation where only the time part is discretized.

We will first study the semi-discrete schemes, then discuss the
discretization of spatial terms of the form $\Div\,\left(a\nabla
  u\right)$ that appear both in the spatial regularizer term and the
tow higher order $\nabla_u E_3^i$, $i=2,3$.
the spatial discretization.  In the inpainting denoising settings, we
need to discretize the gradient of the term $E_1$ as given by
\eqref{eq:graduE1}, this almost trivial, once one has taken care of
multiresolution discretization for $\chi_\Omega$. We thus discuss
briefly points related to the multiresolution setting.  We end by
discussing the approaches we have used to implement the final
numerical solutions of the algebraic equations obtained from
discretization.

In the sequel we will assume that we have a spatio-temporal grid with
spatial grid spacing $h_s$ (we assume the spatial grid to be squared, for
simplicity) and temporal grid spacing $h_t$.
Given a function $f:\RR^2\times\RR\to\RR^k$, its temporal discretization
will give rise to a family 
$$
(f_k)_{k\in\ZZ},\quad f_k = f(-,-,kh_t):\RR^2\to\RR^k.
$$
We will not consider explicitly the full discretization, as the
general need to estimate several quantities at non-grid point
locations would make it very cumbersome.

\subsection{A time-discrete scheme.}
\label{sec:tds}
In order to discretize the terms $\nabla_\vv E_3^i$, we start with a
temporal-only discretization of them, and this for two main reasons. First,
displacement fields have generally subpixel accuracy, necessitating some
spatial interpolation. Second: temporal discrete sequence modeling is at the
heart of multiresolution / warping handling of motion recovery and most of
these warping based motion recovery methods all use at some point the
Lie Derivative semi-discrete approximation in formula \eqref{eq:liederfwrd}.

We will indicate some of the differences that naturally appear in the
gradient terms with respect to flow in this setting.

In order to develop schemes, we start from the continuous energy terms that
we discretize in time and derive analogues of Prop \ref{prop:lieadj}, using
the Lie derivative forward difference expression defined in formula
\eqref{eq:liederfwrd}. To get rid of the boundary conditions difficulties
that arise from boundary integral in Prop \ref{prop:lieadj}, they would be
much worse in the semi-discrete case, we extend artificially the spatial
domain to the whole of $\RR^2$ and the temporal range to the whole of $\RR$,
while we will assume that both the sequence and the velocity field have
compact support.

Let $\phi, \psi:\RR^2\times\RR\to\RR^k$ be sequence, $(\phi_k)_{k\in\ZZ}$
and $(\psi_k)_{k\in\ZZ}$ their temporal discretizations. Denote by
$\bv_k:\RR^2\to\RR^2$ the displacement field $\bv(-,kh_t,h_t)$ where
$\bv(\bx,t,h$) was defined in formula \eqref{eq:dispfield}.  Following
\eqref{eq:liederfwrd} and \eqref{eq:liederbwrd}, we set, with some abuse of
notation, and dropping the subscript $t$ from the temporal spacing $h_t$,
$$
\left(\LieDFh{\bv}{\phi}\right)_k :=
\frac{\phi_{k+1}(\bx+\bv_k(\bx))-\phi_k(\bx)}{h},\quad
\left(\LieDBh{\bv}{\phi}\right)_k :=
\frac{\phi_k(\bx) - \phi_{k-1}(\bx-\bv_k(\bx))}{h},\quad k\in\ZZ.
$$
For $k\in\ZZ$, set $H_k = I + \bv_k$, where $I$ is the identity map of
$\RR^2$. For $h$ small enough, this is a local diffeomorphism if the
velocity field $v$ is smooth. Because of the compact support assumption, $v$
is bounded in norm, implying that for $h$ small enough, $H_k$ is a
diffeomorphism. 
\begin{prop}
  \label{prop:disclieadj}
  Assume that the $H_k$'s are global diffeomorphisms. Define $(K_k)_k$
  by $K_{k+1}$ = $H_{k}^{-1}$, and set $\bw_k = I- K_k$. If $|JK_k|$ is the
  Jacobian determinant of $K_k$, $|JK_k| = 1 - \Div\,\bw_k + |J\bw_k|$ and
  the following holds:
  $$
  \sum_{k\in\ZZ}\int_{\RR^2}\left(\LieDFh{\bv}{\phi}\right)_k(\bx)\psi_k(\bx)\,d\bx
  =
  -\sum_{k\in\ZZ}\int_{\RR^2}\phi_k\left(\left(\LieDBh{\bw}{\psi}\right)_k 
    -\frac{\Div\,\bw_k - |J\bw_k|}{h}\psi_{k-1}\circ K_k\right)\,d\bx.
  $$
\end{prop}
The proof is straightforward via the change of variables theorem and
renaming of summation and integration variables. The \emph{backward
  displacement field} $\bw_k$ is given by
$$
\bw_k(\bx) = -\int_0^{h} \vv(\bx(kh - \tau),kh-\tau)\,d\tau.
$$
and passing to the limit $h\to 0$, $kh\to t_0$ in the above proposition, one easily gets
Prop.~\ref{prop:lieadj} (it is easy to check that in the limit $\Div\,\bw_k/h\to
-\Div\,\vv$ and $|J\bw_k|/h\to 0$) with the same problems in term of modeling: $\vv$ is
generally not smooth: it should present some discontinuities, due to
occlusion/disocclusion. Here too, theoretical and practical difficulties appear, and we do
not consider this case, a practical solution is partially provided by the use of non
quadratic data and regularizing terms in the variational formulations, and some smoothness
can anyway be justified by the smoothing effect of the discretization.

Note that the result stated in Prop.~\ref{prop:disclieadj} is essentially an
adjunction result for the Hilbert space of functions families with inner product
$$
\left(\bphi|\bpsi\right) = \sum_{k\in\ZZ}\int_{\RR^2}\phi_k\psi_k\,d\bx
$$
with the associated notion of gradient that we use in the sequel.

We will use the displacement fields and warp notations from
Prop. \ref{prop:disclieadj}  and as we had introduced in Subsection
\ref{sec:tpmlt}, the spatio-temporal coefficient functions $B_i$,
$i=1,2,3$, we introduce their counterparts sequences $B_i^{\pm}$,
$i=1,2,3$, respectively defined by:
$$
B_{1k}^+ = \phi'\left(\left(\LieDFh{\bv}{u}\right)^2_k\right),\quad
B_{1k}^- = \phi'\left(\left(\LieDBh{\bw}{u}\right)^2_k\right),
$$
$$
B_{2k}^+ = \phi'\left(\left(\LieDFh{\bv}{u}\right)_k^2 + 
  \gamma\left|\LieDFh{\bv}{\nabla u}\right|^2_k\right),\quad
B_{2k}^- = \phi'\left(\left(\LieDBh{\bw}{u}\right)_k^2 + 
  \gamma\left|\LieDBh{\bw}{\nabla u}\right|^2_k\right),
\quad\text{and}
$$
$$
B_{3k}^+ = \phi'\left(\left|\LieDFh{\bv}{\nabla u}\right|^2_k\right),\quad
B_{3k}^- = \phi'\left(\left|\LieDBh{\bw}{\nabla u}\right|^2_k\right).
$$
We will also set 
\begin{equation}
\label{eq:fk}
F_k = \frac{\Div\,\bw_k - |J\bw_k|}{h}.
\end{equation}

We look at the simplest term $E_3^1$ and detail the derivation of its
semi-discrete gradient. To be in the assumptions of the above
proposition, we extend the spatial domain to the whole of $\RR^2$ and
assume that the both the image sequence and velocity fields have
compact support. Let $\buu = (u_k)_k$ the family of image frames.
Using forward difference approximation of Lie derivative, the
semi-discretization of $E_3^1$ is
$$
\bar{E}_3^1(\buu,\bv) = \frac{1}{2}\sum_{k\in\ZZ}\int_{\RR^2}\phi
\left(
\left(\LieDFh{\bv}{u}\right)_k^2
\right)\,d\bx.
$$
In order to compute its gradient with respect to $\buu$, we use the
directional derivative approach, and for $\bxi = (\xi_k)_k$, we
compute $d\bar{E}_{\buu3}^1\bxi = \ell'(0)$ where $\ell(\tau) =
\bar{E}_3^1(\buu+\tau\bxi,\bv)$ to obtain
$$
d\bar{E}_{3\buu}^1\bxi = \sum_{k\in\ZZ}\int_{\RR^2} B_{1k}^+
\left(\LieDFh{\bv}{u}\right)_k\left(\LieDFh{\bv}{\xi}\right)_k\,d\bx.
$$
Applying the above proposition, one gets
$$
d\bar{E}_{3\buu}^1\bxi =
-\sum_{k\in\ZZ}\int_{\RR^2}\xi_k\left[\LieDBh{\bw}{\left(B_{1}^
      +\left(\LieDFh{\bv}{\buu}\right)\right)_k}
  -F_k\left(\left(B_{1k-1}^+\left(\LieDFh{\bv}{\buu}\right)_{k-1}\right)\circ
    K_k\right)\right]\,d\bx
$$
from which the $k$-component of the sought gradient can be written as
$$
\left(\nabla_\buu\bar{E}_1^2\right)_k = -\LieDBh{\bw}{\left(B_{1}^+\left(\LieDFh{\bv}{\buu}\right)\right)_k}
    + F_k\left(\left(B_{1k-1}^+\left(\LieDFh{\bv}{\buu}\right)_{k-1}\right)\circ K_k\right)
$$ 
But, by a straightforward calculation, since $K_k = H_{k-1}^{-1}$, one gets that
$$
\left(\LieDFh{\bv}{\buu}\right)_{k-1}\circ K_k = \left(\LieDBh{\bw}{\buu}\right)_k, \quad
B^+_{ik-1}\circ K_k = B^-_{ik}.
$$
For a given $\bx \in \RR^2$, set $\bx_k^+ = H_k(\bx) = \bx +
\bv_k(\bx)$, $\bx_k^- = K_k(\bx) = \bx - \bw_k(\bx)$. Then we can
write
\begin{subeqnarray}
  \label{eq:disce31grad}
  \slabel{eq:disce13udiff}
  \left(\nabla_\buu\bar{E}_3^1\right)_k(\bx) &=&
  -\frac{B_{1k}^- u_{k-1}(\bx^-_k) -(B_{1k}^- + B_{1k}^+)u_k(\bx) + B_{1k}^+ u_{k+1}(\bx^+_k)}{h^2}\\
   \slabel{eq:disce13utrans}
  &+&B_{1k}^-\frac{u_k(\bx) - u_{k-1}(\bx_k^-)}{h}F_k.
\end{subeqnarray}
The spatial grid of the problem is usually imposed, and a spatial
interpolation technique is necessary to compute the $u(\bx_k^{\pm})$
and the $B_i^{\pm}$. In this work, we have used bilinear and bicubic
interpolations.  A natural question, when implementing schemes based
on the above partial discretization is what to do when $\bx_k^{\pm}$
falls out of the numerical domain.  The easiest solution is then use
the value at $\bx$ instead, this means that we impose
$(\LieDFh{\bv}{\buu})_k = 0$ (resp. $(\LieDBh{\bw}{\buu})_k = 0$) which
are the numerical translations of the boundary conditions we discussed
in Paragraph \ref{sec:bconds}.
The same type of operations are applied for the semi-discrete version of
energy $E_3^2$
$$
\bar{E}_3^2(\buu,\bv) = \frac{1}{2}\sum_{k\in\ZZ}\int_{\RR^2}
\phi\left(\left(\LieDFh{\bv}{\buu}\right)_k^2 + 
  \gamma\left|\left(\LieDFh{\bv}{\nabla \buu}\right)_k\right|^2\right)\,d\bx.
$$
This lead to the following gradient
\begin{subeqnarray}
  \label{eq:disce32grad}
  \left(\nabla_u\bar{E}_3^2\right)_k &=& 
  -\frac{B_{2k}^- u_{k-1}(\bx^-_k) -(B_{2k}^- + B_{2k}^+)u_k(\bx) + B_{2k}^+ u_{k+1}(\bx^+_k)}{h^2}\\
  &+&B_{2k}^-\frac{u_k(\bx) - u_{k-1}(\bx_k^-)}{h}F_k\\
  &+&\gamma\Div\,\left(
    \frac{B_{2k}^- \nabla u_{k-1}(\bx^-_k) -
      (B_{2k}^- + B_{2k}^+)\nabla u_k(\bx) + 
      B_{2k}^+ \nabla u_{k+1}(\bx^+_k)}{h^2}
  \right)\\
  &-&\gamma\Div\,\left(B_{2k}^-\frac{\nabla u_k(\bx)-\nabla u_{k-1}(\bx_k^-)}{h}F_k\right)  
\end{subeqnarray}
while, for energy
$$
\bar{E}_3^3(\buu,\bv) = \frac{1}{2}\sum_{k\in\ZZ}\int_{\RR^2}\left[
\phi\left(\left(\LieDFh{\bv}{\buu}\right)_k^2\right) + 
  \gamma\phi\left(\left|\left(\LieDFh{\bv}{\buu}\right)_k\right|^2\right)
\right]\,d\bx
$$
the gradient is given by
\begin{subeqnarray}
  \label{eq:disce33grad}
  \left(\nabla_u\bar{E}_3^3\right)_k &=& 
  -\frac{B_{1k}^- u_{k-1}(\bx^-_k) -(B_{1k}^- + B_{1k}^+)u_k(\bx) + B_{1k}^+ u_{k+1}(\bx^+_k)}{h^2}\\
  &+&B_{1k}^-\frac{u_k(\bx) - u_{k-1}(\bx_k^-)}{h}F_k\\
  &+&\gamma\Div\,\left(
    \frac{B_{3k}^- \nabla u_{k-1}(\bx^-_k) -
      (B_{3k}^- + B_{3k}^+)\nabla u_k(\bx) + 
      B_{3k}^+ \nabla u_{k+1}(\bx^+_k)}{h^2}
  \right)\\
  &-&\gamma\Div\,\left(B_{3k}^-\frac{\nabla u_k(\bx)-\nabla u_{k-1}(\bx_k^-)}{h}F_k\right). 
\end{subeqnarray}
In the three cases care must be taken when dealing with the
transport-like terms that appear in the different expressions above.
While schemes resulting from \eqref{eq:disce31grad} are relatively
easy to implement, modulo of course transport-like term, care is to be
taken for schemes \eqref{eq:disce32grad} and \eqref{eq:disce33grad}
due to the presence of spatial divergence. When properly dealt with, a
numerical scheme is then available.

Before continuing, important points regarding the above energies and
schemes must be addressed.
\begin{itemize}
\item The use of the forward difference approximations in the energies
  have also an effect on their gradients with respect to the
  displacement field variable. An elementary computation for the case
  of $\bar{E}_3^1$ gives
  $$
  \left(\nabla_\bv\bar{E}_3^1\right)_k = B_1k^+\left(\LieDFh{\bv}{\buu}\right)_k\nabla u_{k+1}\circ H_k
  $$
  where the warping $H_k$ appears naturally and corresponds precisely
  to the type of computations performed in multiresolution motion
  recovery. Remark that only $H_k$ is used there and that there is no
  assumption on its inversibility.
\item Being coherent with Lie derivatives approximations above, our
  motion recovery algorithm will normally return only the forward
  displacement field $\bv=(\bv_k)_k$ of the sequence, and thus we have
  only directly access to the forward warp $H_k$. How to compute the
  backward displacement field $\bw$? Inversion of the $H_k$ is
  generally not an option: even in the case where $H_k$ was a true
  diffeomorphism, this would extremely complex.  Moreover occlusion
  and disocclusion phenomena make it impossible. However, a simple
  solution consists in computing this backward flow from the image
  sequence itself. This should provide a reasonable solution for at
  least a good reason: Given an image $u:\RR^2\times[0,T]\to \RR$, let
  us denote by $\hat{u}$ the ``time-reversed'' sequence obtained as
  $$
  \hat{u}(\bx,t) := u(\bx,T-t).
  $$
  Assume that $u$ is \emph{smooth}. Then, taking the flow related part
  of our energy, i.e. $E_{ij} = \lambda_3E_3^i + \lambda_4 E_4^j$, an
  elementary computation shows that if $v$ minimizes $E_{ij}$, so is
  $-v$ for the image sequence $\hat{u}$. We could replace the optical
  flow estimation by a symmetrized version, in the spirit of the work
  of Alvarez \etal in \cite{alvarez-deriche-etal:02}, but this would
  deeply modify the formulations above.
\item All the above gradients have some transport-like terms, e.g the
  term \eqref{eq:disce13utrans} in the expression of
  $\nabla_\buu\bar{E}_3^1$. Such a term may be difficult to handle
  numerically. One can ask whether it is necessary. In an informal
  way, we do expect that the diffusion term \eqref{eq:disce13udiff}
  alone will smooth variations of $u$ along the flow lines, and thus
  decrease the energy, i.e. that even when forgetting the transport
  term, the resulting expression would still be a descent direction
  for the energy, so should be consider for both explicit gradient
  descent resolution as well as some relaxation schemes. This however
  no so simple, as \eqref{eq:disce13udiff} involves not only forward
  Lie derivatives but also backward ones while only forward ones are
  present in the semi-discrete energy. In the other hand, our flow
  regularizers generally favor displacement fields with small
  divergence, thus generally reducing the influence of the transport
  part.
\end{itemize}

\subsection{Multiresolution details}
\label{sec:mrdet}

As mentioned in Section \ref{sec:mres}, we solve the equations in a
multiresolution framework. A few points should be mentioned. When
building the multiresolution pyramid, image coarsening is necessary,
and is in fact already handled by the motion estimation solver. What
is not handled is the coarsening of the inpainting mask, i.e the
numeric characteristic function of the missing data locus $\Omega$.
Coarsening of the image is usually performed by smoothing ans
subsampling or any kind of method that has a proper low pass property.
Coarsening of $\chi_\Omega$ may be problematic when the same method is
used.  It blurs the boundary of $\Omega$, and while this blurring may
seem coherent, as it may indicate that a given pixel at coarse
resolution contains some known partial information, we have constated
that it often slows convergence down.  We have used instead the
somehow rough but simple nearest neighbor approach that guaranties that
the mask remains binary valued and not become ``too large'' when
coarsening.

In the other hand, one also need to interpolate the inpainting result
from a given level to the next finer level, while interpolation of the
motion is performed already by the flow recovery algorithm. In this
work, image interpolation has been performed using a simple bilinear
interpolation for inpainting denoising, while for pure inpainting,
values obtained by bilinear interpolation and not in the missing data
locus have been replaced by original image values downsampled at that
resolution.

\subsection{Solving the equations}
\label{sec:solv}

Inpainting equations involving terms $E_3^2$ and $E_3^3$ are 4th-orders
partial differential equations, with diffusion of gradients. This type
of higher order diffusion may not obey the minimum-maximum principle,
see for instance \cite{gilboa-zeevi-etal:04}. We rely therefore on a
gradient descent scheme in these cases. For each resolution level we
do the following.  Having chosen an evolution step $d\tau$, we create
the family $\buu^n = (u_k^n)_k$, $n\geq 0$ and write a standard
Eulerian step
$$
\frac{u^{n+1}_k(\bx) - u^n_k(\bx)}{d\tau} = -\left(\nabla_{\buu^n}
  \bar{E}(\bx)\right)_k
$$ 
where $\bar E$ is either of the form $\lambda_1 \bar{E_1} +
\lambda_2\bar{E_2} + \lambda_3 E_3^{i}$ in the inpainting-denoising
case, with $(\bx,k)$ running over the full spatio-temporal grid in
that case, while in the pure inpainting case $\bar{E}$ has the form
$\lambda_2\bar{E_2} + \lambda_3 E_3^{i}$ and $(\bx,k)$ runs only on
the missing data locus, i.e the discretized and downsampled copy of
$\Omega$. 

In the implementation, we choose once for all the evolution step
$d\tau$ as well as the number $N$ of evolution steps we perform. As in
most explicit schemes, $d\tau$ must be chosen small enough, and this
has a drastic impact on the running-time, particularly in the case of
inpainting denoising.

In the case where we use $E_3^1$, we may consider, in a fixed point
approach, linearizing the resulting equations and solvers such as
Gauss-Seidel could be used. In order to be able to compare the
different methods, we have however in that work only used the explicit
Gradient Descent approach.

\section{Experimental Evaluation}
\label{sec:results}

We present results for several of the algorithms we have discussed, on
synthetic and real sequences. We follow the nomenclature of energy
terms presented in Section \ref{sec:mcbayes-var}, Tables
\ref{tab:intmot} and \ref{tab:motreg}.  To accommodate the variety of
intensity ranges in the different sequences, intensities have been
linearly normalized to range $[0,1]$. This imply that numerical
partial derivatives are bounded in absolute value, by bounds of the
same order of magnitude, which in turns influences the practical range
of $\varepsilon$ values in $\phi(x^2) = \sqrt{x^2 + \varepsilon^2}$,
we have taken $\varepsilon = 10^{-3}$. In all gradient descents, we
have used an algorithmic time step $dt = 10^{-3}$. This in turns
influences the choices of the different weights $\lambda_i$ for the
corresponding energies.

Although flow regularization term $E_4^1$ has been used, we only
report results for term $E_4^2$ in the experiments we present here, as
it has been argued \cite{brox-bruhn-etal:04,bruhn-weickert:05} that
temporal regularity prior generally improves motion estimation.

We present a series of stills in this section, the reader
should also look at the companion sequences, they are available at the
location \verb|http://image.diku.dk/francois/seqinp|. Some previous
versions were available at the the companion web site
\verb|http://www-sop.inria.fr/books/imath|  for the
monograph of Aubert and Kornprobst \cite{aubert-kornprobst:06}.

The first one is the well known \texttt{Yosemite} sequence, created by
Lynn Quam, very often used for optical flow evaluation. The sequence
is artificial and the ground truth is known for the flow. It was
degraded by removing large polygonal patches, three of them
\emph{overlapping consecutively} in time, on 6 of the 15 frames.

\begin{figure}[htpb]
  \centerline{
    \ifthenelse{\boolean{PNG}}
    {
      \begin{tabular}{c@{\hspace*{1mm}}c@{\hspace*{1mm}}c}
        \includegraphics[width = \three]{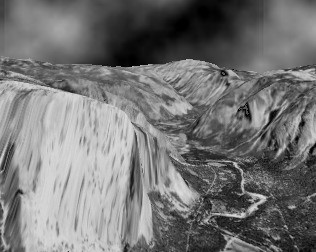}&
        \includegraphics[width = \three]{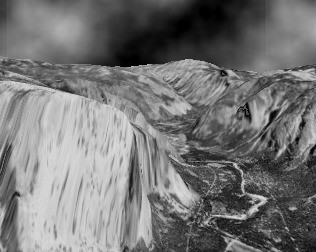}&
        \includegraphics[width = \three]{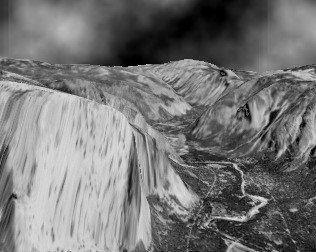}\\
        (a) original frame 3 & (b) original frame 4 & (c) original frame 5\\
        \includegraphics[width = \three]{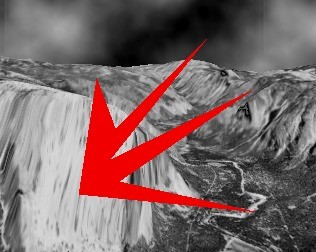}&
        \includegraphics[width = \three]{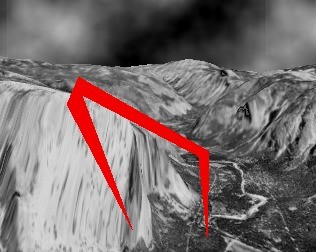}&
        \includegraphics[width = \three]{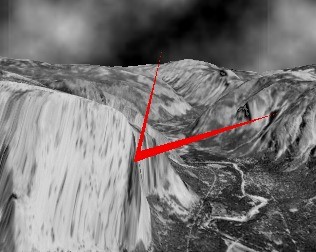}\\
        (d) degraded frame 3 & (e) degraded frame 4 & (d) degraded frame 5 \\
        \includegraphics[width = \three]{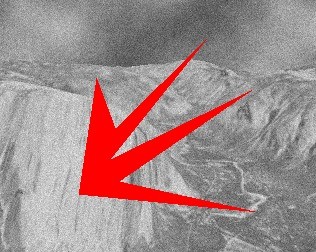}&
        \includegraphics[width = \three]{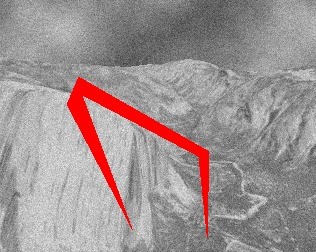}&
        \includegraphics[width = \three]{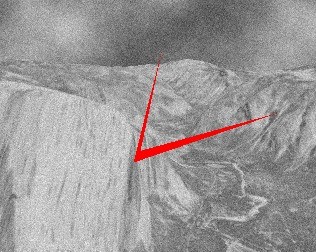}\\
      \end{tabular}
    }
    {
      \begin{tabular}{c@{\hspace*{1mm}}c@{\hspace*{1mm}}c}
        \includegraphics[width = \three]{images/yos/orig/yosorig_02.eps}&
        \includegraphics[width = \three]{images/yos/orig/yosorig_03.eps}&
        \includegraphics[width = \three]{images/yos/orig/yosorig_04.eps}\\
        (a) original frame 3 & (b) original frame 4 & (c) original frame 5\\
        \includegraphics[width = \three]{images/yos/degradedred/yosdeg_02.eps}&
        \includegraphics[width = \three]{images/yos/degradedred/yosdeg_03.eps}&
        \includegraphics[width = \three]{images/yos/degradedred/yosdeg_04.eps}\\
        (d) degraded frame 3 & (e) degraded frame 4 & (d) degraded frame 5 \\
        \includegraphics[width = \three]{images/yos/degradedredgn/yosnoise_02.eps}&
        \includegraphics[width = \three]{images/yos/degradedredgn/yosnoise_03.eps}&
        \includegraphics[width = \three]{images/yos/degradedredgn/yosnoise_04.eps}\\
      \end{tabular}
    }
  }
  \caption{\texttt{Yosemite} sequence. Original and degraded
    frames. The degradation in frame 3 is very large and one can
    notice that the 3 holes overlap around the frame centers.}
  \label{fig:yosframes}
\end{figure}

Figure \ref{fig:yosframes} shows frames 2, 3, and 4 of the original
sequence, corresponding degraded frames, and the noisy degraded ones
where Gaussian noise of standard deviation 5\% of the intensity range
was added.

We first run a simplified experiment where only the pure inpainting
equation derived from terms $E_2$ and $E_3^1$ is solved and then the
inpainting/denoising using the same spatial and trajectory smoothness
terms. We use the ground truth forward flow $\vv_f$. We deal with
the absence of ground truth for the backward optical flow $\vv_b$
by computing it from the optical flow PDE derived from terms $E_3^1$
and $E_4^2$, having reversed time in the sequence and using as
starting guess $t\mapsto-\vec{v}_f(T-t)$ where $T$ is the forward flow
sequence last time/frame. 

\begin{figure}[htbp]
  \centerline{
    \ifthenelse{\boolean{PNG}}
    {
      \begin{tabular}{c@{\hspace*{1mm}}c@{\hspace*{1mm}}c}
        \includegraphics[width = \three]{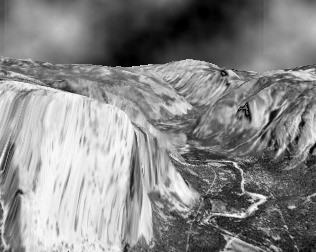}&
        \includegraphics[width = \three]{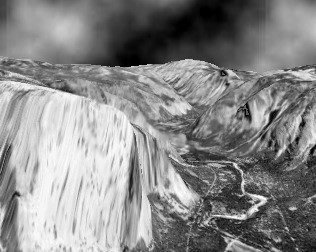}&
        \includegraphics[width = \three]{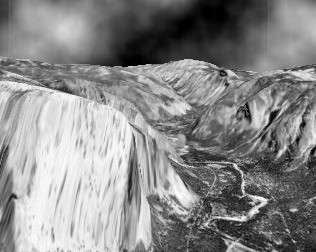}\\
        (a) pure inpainting frame 3 & (b) pure inpainting frame 4 &(c) pure inpainting frame 4\\
        \includegraphics[width = \three]{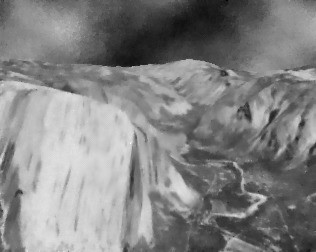}&
        \includegraphics[width = \three]{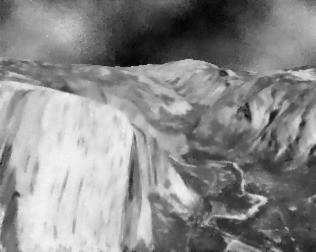}&
        \includegraphics[width = \three]{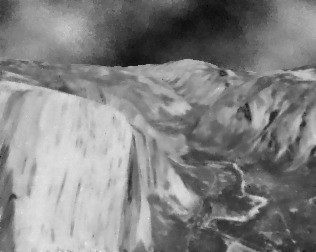}\\
        (d) inpainted/denoised frame 3 & (e) inpainted/denoised frame 4 &(f) inpainted/denoised frame 5
      \end{tabular}
    }
    {
      \begin{tabular}{c@{\hspace*{1mm}}c@{\hspace*{1mm}}c}
        \includegraphics[width = \three]{images/yos/pureinpgtm/yospinm_02.eps}&
        \includegraphics[width = \three]{images/yos/pureinpgtm/yospinm_03.eps}&
        \includegraphics[width = \three]{images/yos/pureinpgtm/yospinm_04.eps}\\
        (a) pure inpainting frame 3 & (b) pure inpainting frame 4 &(c) pure inpainting frame 4\\
        \includegraphics[width = \three]{images/yos/idgtm/yosidgtm_02.eps}&
        \includegraphics[width = \three]{images/yos/idgtm/yosidgtm_03.eps}&
        \includegraphics[width = \three]{images/yos/idgtm/yosidgtm_04.eps}\\
        (d) inpainted/denoised frame 3 & (e) inpainted/denoised frame 4 &(f) inpainted/denoised frame 5
      \end{tabular}
    }
  }
  \caption{Pure Inpainting and Inpainting/Denoising based on ground
    truth motion. While pure inpainted sequence is of excellent
    quality, the inpainted/denoised shows the usual patterns of
    Total-Variation like regularization with loss of fine scale
    details.}
  \label{fig:yosexp12}
\end{figure}

Figure \ref{fig:yosexp12} present the results of these two experiments
on the three degraded frames show in Fig. \ref{fig:yosframes}. For
pure Inpainting, parameters where $\lambda_2 = 0.1$ (spatial
regularization weight), $\lambda_3 = 1$ (flow lines regularization
weight). For inpainting / denoising, the data weight $\lambda_1$ has
been set to 20, while $\lambda_2=\lambda3 = 1$. Results of pure
inpainting presented in (a), (b) and (c) are generally very good, one
may however notice a lack on sharpness in the center of frames (a) and
(b).  Unsurprisingly, inpainted/denoised results present characteristics
of Total Variation regularized images, some low scale details have
been lost. Boulanger \etal \cite{boulanger-kervrann-bouthemy:07} as
well as Buades \etal \cite{buades-coll-morel:08} have shown that image
sequence denoising is best achieved with patch based / non local means
methods. Nevertheless, they cannot cope as is with large missing data.
It may be worth investigating developing an hybrid method in that case.

\begin{figure}[htbp]
  \centerline
  {
    \ifthenelse{\boolean{PNG}}
    {
      \begin{tabular}{c@{\hspace*{1mm}}c@{\hspace*{1mm}}c}
        \includegraphics[width = \three]{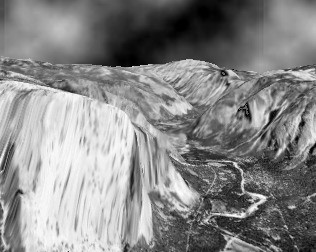}&
        \includegraphics[width = \three]{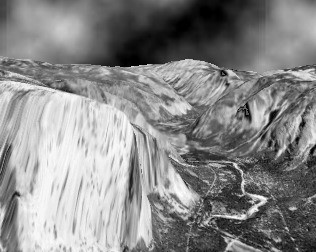}&
        \includegraphics[width = \three]{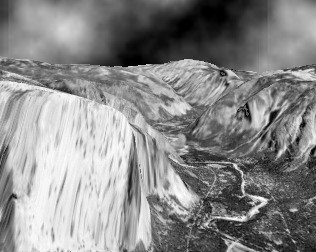}\\
        (a) & (b)  &(c) \\
        \includegraphics[width = \three]{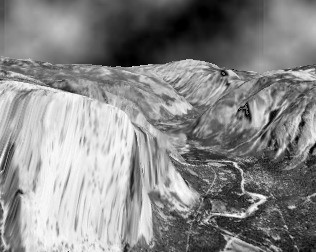}&
        \includegraphics[width = \three]{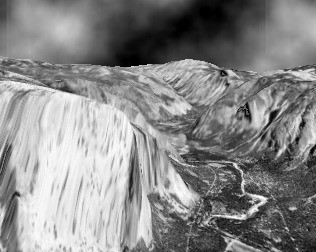}&
        \includegraphics[width = \three]{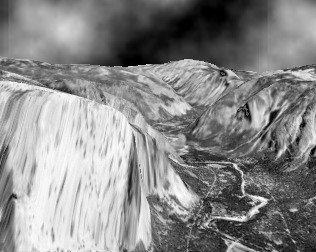}\\
        (d)  & (e)  &(f) \\
        \includegraphics[width = \three]{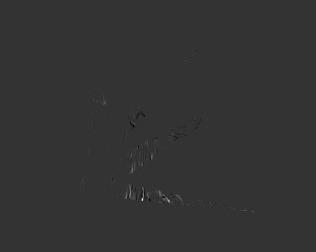}&
        \includegraphics[width = \three]{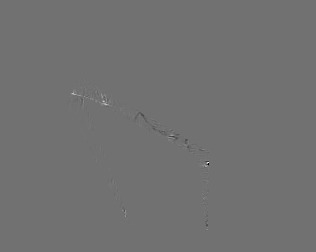}&
        \includegraphics[width = \three]{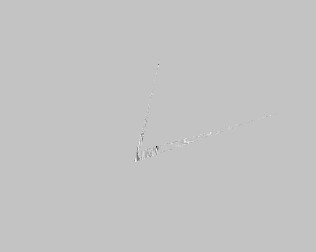}\\
        (g) & (h) & (i)
      \end{tabular}
    }
    {
      \begin{tabular}{c@{\hspace*{1mm}}c@{\hspace*{1mm}}c}
        \includegraphics[width = \three]{images/yos/pimce31/yosmce2e31_02.eps}&
        \includegraphics[width = \three]{images/yos/pimce31/yosmce2e31_03.eps}&
        \includegraphics[width = \three]{images/yos/pimce31/yosmce2e31_04.eps}\\
        (a) & (b)  &(c) \\
        \includegraphics[width = \three]{images/yos/pimce32/yosmce2e32_02.eps}&
        \includegraphics[width = \three]{images/yos/pimce32/yosmce2e32_03.eps}&
        \includegraphics[width = \three]{images/yos/pimce32/yosmce2e32_04.eps}\\
        (d)  & (e)  &(f) \\
        \includegraphics[width = \three]{images/yos/diffe31e32/diffe31e32_02.eps}&
        \includegraphics[width = \three]{images/yos/diffe31e32/diffe31e32_03.eps}&
        \includegraphics[width = \three]{images/yos/diffe31e32/diffe31e32_04.eps}\\
        (g) & (h) & (i)
      \end{tabular}
    }
  }
  \caption{ Full Image and Motion recovery in the case of pure Inpainting. The first row
    from minimization of low-order energy $E_2+E_3^1+E_4^2$, the second from higher order
    energy $E_2+E_3^2+E_4^2$ and the last row shows the differences.}
  \label{fig:yosmcpi}
\end{figure}

\begin{figure}[htbp]
  \centerline
  {
    \ifthenelse{\boolean{PNG}}
    {
      \begin{tabular}{c@{\hspace*{1mm}}c}
        \includegraphics[width = \two]{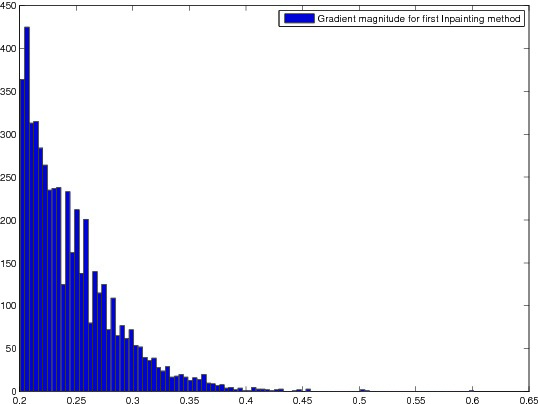} &  
        \includegraphics[width = \two]{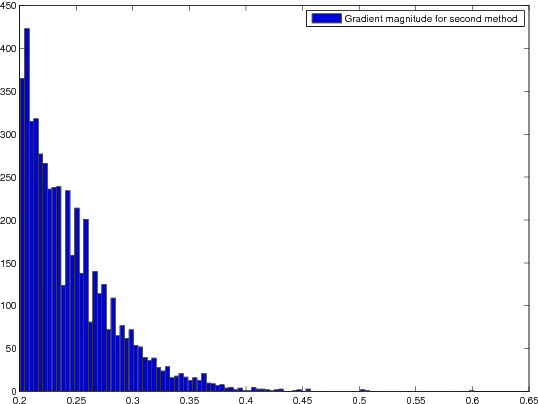}
      \end{tabular}
    }
    {
      \begin{tabular}{c@{\hspace*{1mm}}c}
        \includegraphics[width = \two]{images/gmage31.eps} &  
        \includegraphics[width = \two]{images/gmage32.eps}
      \end{tabular}
    }
  }
  \caption{Comparisons of gradient magnitudes in inpainted regions. (a) when using energy
    term $E_3^1$, (b) when using energy term $E_3^2$. They are almost identical.}
  \label{fig:gradhisto}
\end{figure}

In the second set of experiments, we abandon the ground truth and run
full recovery algorithms on the \texttt{Yosemite} sequence.  In the
first one, we run the pure inpainting algorithm corresponding to
energy $\lambda_2 E_2 + \lambda_3 E_3^1 + \lambda_4 E_4^2$ with
$\lambda_2 = 0.1$, $\lambda_3 = 1$ and $\lambda_4 = 0.2$, four pyramid
levels where the number of pixels is roughly divided by 2 from one
level to the next coarser one. Then Energy $\lambda_2 E_2 + \lambda_3
E_3^1 + \lambda_4 E_4^2$ has been used, withe the gradient weight
$\gamma = 0.1$, and \emph{these parameters, as well as pyramid sizes,
  have been used in all the remaining experiments presented in this
  work, except for the \texttt{Manon} sequence}. Results are shown in
Figure \ref{fig:yosmcpi}. Spotting visually differences between the
two sequences is difficult, although computing the differences, as
illustrated on the last row of Figure \ref{fig:yosmcpi} shows some,
covering about 7\% of the intensity range, while a plot of histograms
of gradient magnitudes for the two results seems to indicate that this
difference is in fact hardly significative. These histograms are shown
in figure \ref{fig:gradhisto}.

\begin{figure}[htbp]
  \centerline{
    \ifthenelse{\boolean{PNG}}
    {
      \begin{tabular}{c@{\hspace*{1mm}}c@{\hspace*{1mm}}c}
        \includegraphics[width = \three]{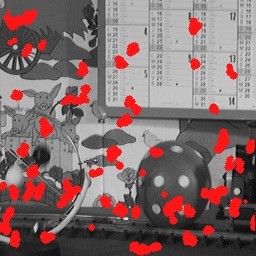}&
        \includegraphics[width = \three]{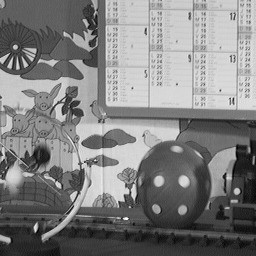}&
        \includegraphics[width = \three]{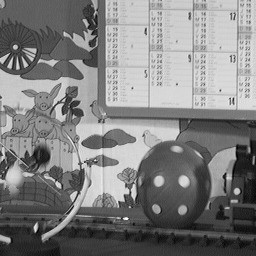}\\
        \includegraphics[width = \three]{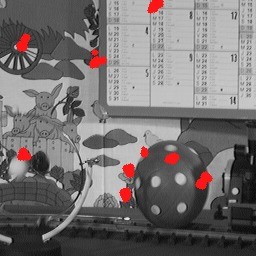}&
        \includegraphics[width = \three]{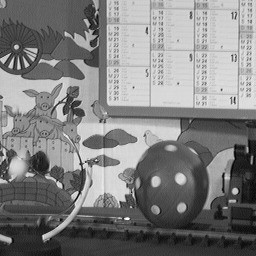}& 
        \includegraphics[width = \three]{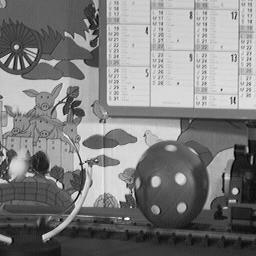}\\
        \includegraphics[width = \three]{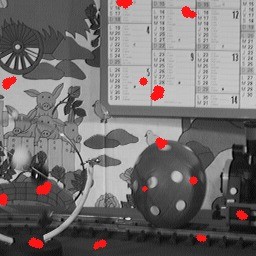}&
        \includegraphics[width = \three]{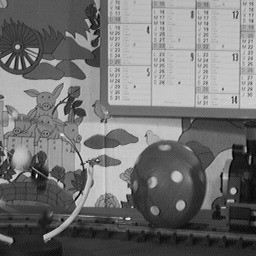}& 
        \includegraphics[width = \three]{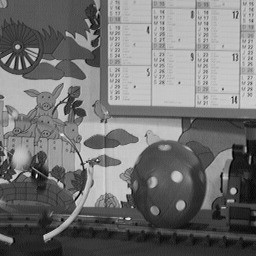}\\
        \includegraphics[width = \three]{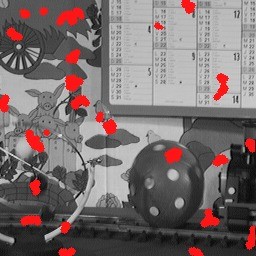}&
        \includegraphics[width = \three]{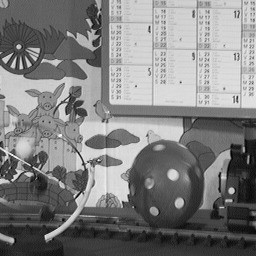}& 
        \includegraphics[width = \three]{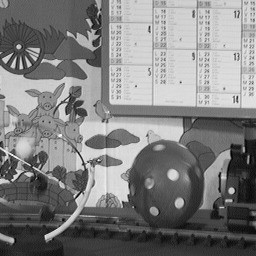}\\
      \end{tabular}
    }
    {
      \begin{tabular}{c@{\hspace*{1mm}}c@{\hspace*{1mm}}c}
        \includegraphics[width = \three]{images/mobcal/orig/mobdeg_10.eps}&
        \includegraphics[width = \three]{images/mobcal/inpvar/mobvar_10.eps}&
        \includegraphics[width = \three]{images/mobcal/inpnvar/mobnvar_10.eps}\\
        \includegraphics[width = \three]{images/mobcal/orig/mobdeg_11.eps}&
        \includegraphics[width = \three]{images/mobcal/inpvar/mobvar_11.eps}& 
        \includegraphics[width = \three]{images/mobcal/inpnvar/mobnvar_11.eps}\\
        \includegraphics[width = \three]{images/mobcal/orig/mobdeg_12.eps}&
        \includegraphics[width = \three]{images/mobcal/inpvar/mobvar_12.eps}& 
        \includegraphics[width = \three]{images/mobcal/inpnvar/mobnvar_12.eps}\\
        \includegraphics[width = \three]{images/mobcal/orig/mobdeg_13.eps}&
        \includegraphics[width = \three]{images/mobcal/inpvar/mobvar_13.eps}& 
        \includegraphics[width = \three]{images/mobcal/inpnvar/mobnvar_13.eps}\\
      \end{tabular}
    }  
  }
  \caption{On the first column, frames 10, 11, 12 and 13 of the degraded
    \texttt{Mobile and Calendar} sequence.  The second column shows
    the results of minimization of energy $E_2 + E_3^3 + E_4^2$ while
    the last column shows the results when the corresponding
    inpainting equation is replaced by the low order one.}
  \label{fig:mobcal}
\end{figure}

The second example is taken from the companion CD ROM of Kokaram's
book~\cite{kokaram:98}. We start with the \texttt{Mobile and Calendar}
sequence, that have been extensively used for MPEG coding,
deinterlacing and so on. It is a real one with 25 frames and with
complex motion patterns. It is then artificially degraded to simulate
blotches. (approximately 6\% of the image is degraded with blocthes of
multiple size, they may overlap in time.  Here too the energy $E_2 +
E_3^3 + E_4^2$ was used, with the same parameters as above.  Figure
\ref{fig:mobcal} presents four frames of the degraded sequence, the
inpainting results using energy $(E_2,E_3^3,E_4^2)$, and a solver
where the optical flow is computed with this energy while for the
inpainting we use the lower order equation (equivalently setting the
gradient parameter $\gamma$ to 0 in $E_3^3$). Differences are non
significative. This too substantiates the idea that the lower order
inpainting equation is a good descent direction.

\begin{figure}[htbp]
  \centerline{
    \ifthenelse{\boolean{PNG}}
    {
      \begin{tabular}{c@{\hspace*{1mm}}c@{\hspace*{1mm}}c}
        \includegraphics[width = \three]{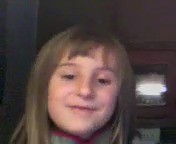}&
        \includegraphics[width = \three]{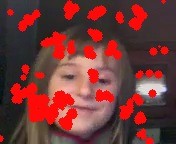}&
        \includegraphics[width = \three]{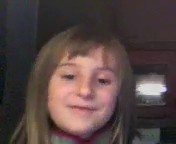}\\
        \includegraphics[width = \three]{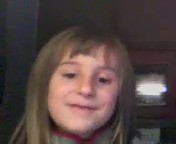}&
        \includegraphics[width = \three]{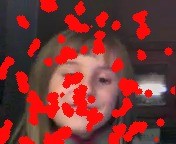}&
        \includegraphics[width = \three]{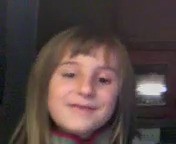}\\
        \includegraphics[width = \three]{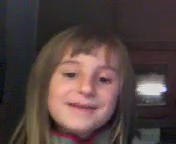}&
        \includegraphics[width = \three]{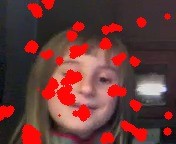}&
        \includegraphics[width = \three]{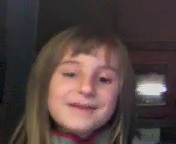}\\
      \end{tabular}
    }
    {
       \begin{tabular}{c@{\hspace*{1mm}}c@{\hspace*{1mm}}c}
        \includegraphics[width = \three]{images/Manon/orig/manon_orig_07.eps}&
        \includegraphics[width = \three]{images/Manon/deg/manon_red_07.eps}&
        \includegraphics[width = \three]{images/Manon/inp/manon_inp_07.eps}\\
        \includegraphics[width = \three]{images/Manon/orig/manon_orig_08.eps}&
        \includegraphics[width = \three]{images/Manon/deg/manon_red_08.eps}&
        \includegraphics[width = \three]{images/Manon/inp/manon_inp_08.eps}\\
        \includegraphics[width = \three]{images/Manon/orig/manon_orig_09.eps}&
        \includegraphics[width = \three]{images/Manon/deg/manon_red_09.eps}&
        \includegraphics[width = \three]{images/Manon/inp/manon_inp_09.eps}\\
      \end{tabular}
    }
  }
  \caption{On the first column, frames 7,8 and 9 of the
    original \texttt{Manon} sequence. The second column shows the
    corresponding degraded frames and the last column the results of
    the color inpainting using Energy $(E_2,E_3^2,E_4^2)$.}
  \label{fig:manon}
\end{figure}

\begin{figure}[htbp]
  \centerline{
    \ifthenelse{\boolean{PNG}}
    {
      \begin{tabular}{c@{\hspace*{1mm}}c@{\hspace*{1mm}}c}
        \includegraphics[width = \three]{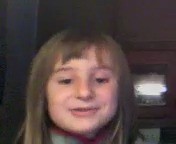}&
        \includegraphics[width = \three]{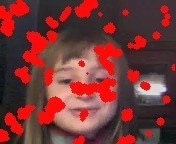}&
        \includegraphics[width = \three]{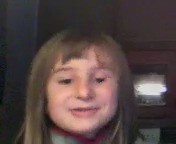}\\
      \end{tabular}
    }
    {
      \begin{tabular}{c@{\hspace*{1mm}}c@{\hspace*{1mm}}c}
        \includegraphics[width = \three]{images/Manon/orig/manon_orig_05.eps}&
        \includegraphics[width = \three]{images/Manon/deg/manon_red_05.eps}&
        \includegraphics[width = \three]{images/Manon/inp/manon_inp_05.eps}\\
      \end{tabular}
    }
  }  
  \caption{Original, degraded and inpainted frame 5 of the
    \texttt{Manon} sequence. Problems can be observed in the
    reconstruction of the left eye as well as part of the hair above
    it.}
  \label{fig:manon5}
\end{figure}

The third sequence, \texttt{Manon} is a real \emph{color} sequence
acquired by the first author with a mobile device, featuring the
author's daughter. The original sequence was encoded with the 3G
format, a simplified MPEG format for third generation mobile
devices. It has been artificially degraded with red blotches and
inpainted with the \emph{vectorial form} of the energy
$(E_2,E_3^2,E_4^2)$ with $\lambda_2 = 0.1, \lambda_3=1, \gamma=0.1$
and $\lambda_4 = 0.02$ a value 5 times smaller than in the other
experiments, because a too large values caused problems in the
recovery of hair motion among others. By its encoding and its nature,
a face with non rigid motion, it presents some challenges. Three
original frames, 7, 8 and 9, the corresponding degraded sequences as
well as the inpainted ones are presented on
Figure~\ref{fig:manon}. The result is visually very
good. Nevertheless, some serious problems were encountered at other
locations of the sequence and Figure \ref{fig:manon5} presents some of
the encountered problems at the 5th frame of the sequence, where a
portion of the hair and of an eye are wrongly interpolated.


\begin{figure}[htbp]
  \centerline{ 
    \ifthenelse{\boolean{PNG}}
    {
      \begin{tabular}{c@{\hspace*{1mm}}c@{\hspace*{1mm}}c}
        \includegraphics[width = \three]{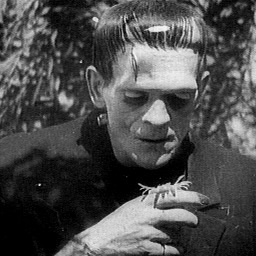}&
        \includegraphics[width = \three]{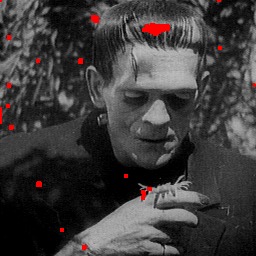}&
        \includegraphics[width = \three]{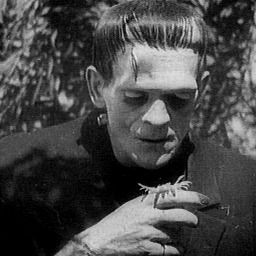}\\
      \end{tabular}
      }
      {
        \begin{tabular}{c@{\hspace*{1mm}}c@{\hspace*{1mm}}c}
          \includegraphics[width = \three]{images/frank/orig/frank_orig_08.eps}&
          \includegraphics[width = \three]{images/frank/detect/frank_detect_08.eps}&
          \includegraphics[width = \three]{images/frank/inp/frank_inp_08.eps}\\
      \end{tabular}
      }
  }  
  \caption{The \texttt{Frankenstein sequence}. From left to right: Frame 8, ROD detected
    defects and inpainting.  by minimizing energy $E_2 + E_3^2 + E_4^2$}
  \label{fig:frank}
\end{figure}

\begin{figure}[htbp]
  \centerline{  
    \ifthenelse{\boolean{PNG}}
    {
      \begin{tabular}{c@{\hspace*{1mm}}c@{\hspace*{1mm}}c}
        \includegraphics[width = \three]{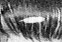}&
        \includegraphics[width = \three]{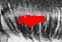}&
        \includegraphics[width = \three]{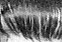}
      \end{tabular}
    }
    {
      \begin{tabular}{c@{\hspace*{1mm}}c@{\hspace*{1mm}}c}
        \includegraphics[width = \three]{images/frank/orig/closup_orig_08.eps}&
        \includegraphics[width = \three]{images/frank/detect/closup_detect_08.eps}&
        \includegraphics[width = \three]{images/frank/inp/closup_inp_08.eps}
      \end{tabular}
    }
  }  
  \caption{The \texttt{Frankenstein sequence}: Close-up of the hair
    blotch, its detection and its inpainting}.
  \label{fig:hair}
\end{figure}

The fourth and last sequence, called \texttt{Frankenstein}, also taken
from Kokaram's book, is a real degraded one, with 64 frames, for which
no ground truth is known. In our experimentations we used only a
subsequence of 21 frames, with frame 8 presenting, among others, a
relatively large blotch on Frankenstein's hair.  In
figure~\ref{fig:frank} we show this frame, the detected blotch and its
reconstruction. In figure~\ref{fig:hair} we show a close-up of the
damaged hair of the character with the detected blotch and the
reconstruction. Blotches were detected using the so called Rank Order
Detector (ROD), as described in \cite{kokaram:98}, modified for the optical
flow algorithm. Fine texture details were very plausibly recreated.



\section{Conclusion}
\label{sec:conclusion}

In this paper we have introduced a generic variational formulation for
joint recovery of motion and intensity in degraded image sequences,
dealing both with noise and missing data. This generic formulation has
been instantiated in several energy formulations, mostly based on
known motion recovery approached. They give rise to system of partial
differential equations, for motion and intensity. We have focused on
the intensity ones and developed schemes to handle them numerically.
We have validated our approach on a series of experiments. While they
provide often excellent results, they are generally computationally
demanding, especially due to higher order equations to solve for
Inpainting.  Is such complex equation necessary? We discussed that a
simpler equation might still provide a good descent direction when
minimizing a higher order energy, and we presented an experiment that
substantiate this idea: there is a difference, but the result obtained
with lower order equation is visually good. The possibility of using a
lower order equation opens the door for much more efficient solvers:
multigrid solvers where developed for optical flow by
\cite{memin-perez:02} as well as \cite{bruhn-etal:06} and we are
currently working on developing fast multigrid schemes for lower
order inpainting equations.  The general multiresolution optimization
framework that we have used, decouples, at each pyramid level, the
computation of the flows and the images. We are also investigating
more coupled methods, also within the multigrid framework.

\bibliographystyle{plain}
\bibliography{bibfrancois}

\end{document}